\documentclass[preprint,12pt]{elsarticle}
\usepackage{lineno,hyperref}
\usepackage{setspace}
\usepackage[caption=false,font=footnotesize,labelfont=rm,textfont=rm]{subfig}
\usepackage{caption}
\usepackage{amsmath}
\usepackage{tabularx}
\newlength{\tblwidth} 
\usepackage{array}
\usepackage{threeparttable}
\usepackage{newfloat}
\usepackage{amssymb} 
\usepackage{graphicx} 
\usepackage{booktabs} 

\usepackage{color}
\usepackage{multirow}
\modulolinenumbers[5]
\usepackage{bbm}
\usepackage{makecell}

\journal{Neurocomputing}

\begin{document}
\begin{frontmatter}
\title{Magic3DSketch: Create Colorful 3D Models From Sketch-Based 3D Modeling Guided by Text and Language-Image Pre-Training}

\author[mymainaddress]{Ying Zang}
\author[mymainaddress]{Yidong Han}
\author[mymainaddress,mythirdaddress]{Chaotao Ding}

\author[mymainaddress]{Jianqi Zhang}
\author[mysecondaryaddress,mythirdaddress]{Tianrun Chen\corref{mycorrespondingauthor}}\cortext[mycorrespondingauthor]{Corresponding author at: College of Computer Science and Technology, Zhejiang University, Hangzhou 310027, China}\ead{tianrun.chen@zju.edu.cn}

\address[mymainaddress]{School of Information Engineering, Huzhou University, Huzhou 313000, China}
\address[mysecondaryaddress]{College of Computer Science and Technology, Zhejiang University, Hangzhou 310027, China}
\address[mythirdaddress]{KOKONI 3D, Moxin (Huzhou) Technology Co., LTD. Huzhou 313000, China}

\begin{abstract}
The requirement for 3D content is growing as AR/VR application emerges. At the same time, 3D modelling is only available for skillful experts, because traditional methods like Computer-Aided Design (CAD) are often too labor-intensive and skill-demanding, making it challenging for novice users. Our proposed method, Magic3DSketch, employs a novel technique that encodes sketches to predict a 3D mesh, guided by text descriptions and leveraging external prior knowledge obtained through text and language-image pre-training. The integration of language-image pre-trained neural networks complements the sparse and ambiguous nature of single-view sketch inputs. Our method is also more useful and offers higher degree of controllability compared to existing text-to-3D approaches, according to our user study. Moreover, Magic3DSketch achieves state-of-the-art performance in both synthetic and real dataset with the capability of producing more detailed structures and realistic shapes with the help of text input. Users are also more satisfied with models obtained by Magic3DSketch according to our user study. Additionally, we are also the first, to our knowledge, add color based on text description to the sketch-derived shapes. By combining sketches and text guidance with the help of language-image pretrained models, our Magic3DSketch can allow novice users to create custom 3D models with minimal effort and maximum creative freedom, with the potential to revolutionize future 3D modeling pipelines. 
\end{abstract}
\begin{keyword}
3D Modeling\sep Shape from X \sep Sketch \sep Text-Prompting
\end{keyword}

\end{frontmatter}

{\begin{figure*}[!h]
\centering
  \includegraphics[width=1\textwidth]{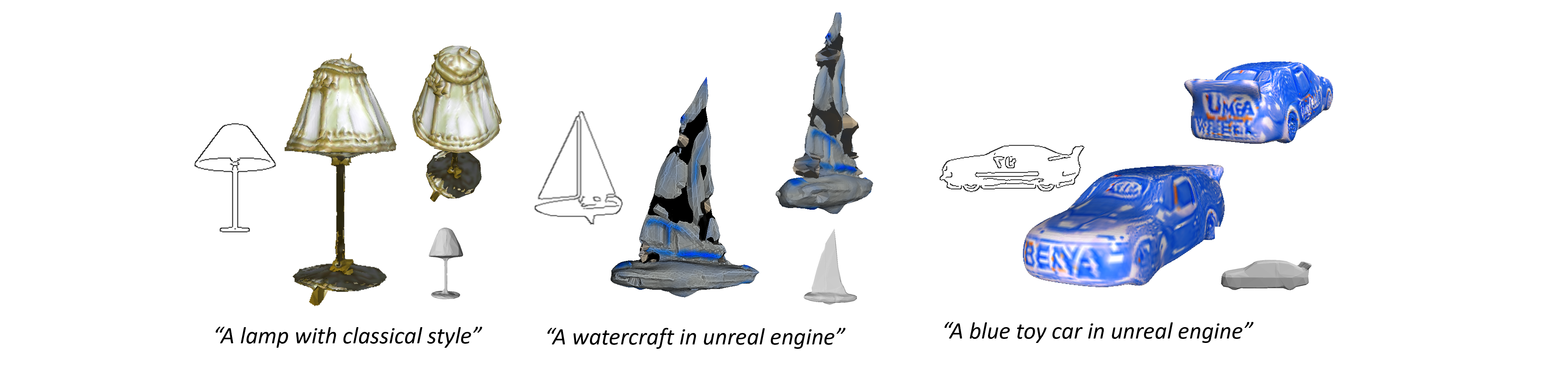}
  \caption{Magic3DSketch utilizes a single sketch as input, along with a text prompt, to generate a high-fidelity and realistic 3D
mesh complete with colors.}
  \label{fig:teaser}
\end{figure*}}


\section{Introduction}

The demand for 3D content has increased significantly in recent years, driven by the rise of virtual and augmented reality technologies and the need for immersive experiences in various fields such as entertainment, education, and product design \cite{wang2020vr}. However, creating professional 3D models is still not for everyone. Wide-adopted 3D modeling approach using Computer-Aided Design (CAD) software is a challenging and time-consuming process that requires a high level of expertise. The CAD software often requires the user to have an extensive knowledge of 3D modeling techniques (\textit{strategic knowledge}), as well as the ability to work with complex interfaces and tools (\textit{command knowledge}) \cite{cohen1999interface}. This creates a barrier for many potential creators who may have great ideas but lack the necessary skills to bring them to life.

{\begin{figure*}[h]
	\centering
	\includegraphics[scale=0.5]{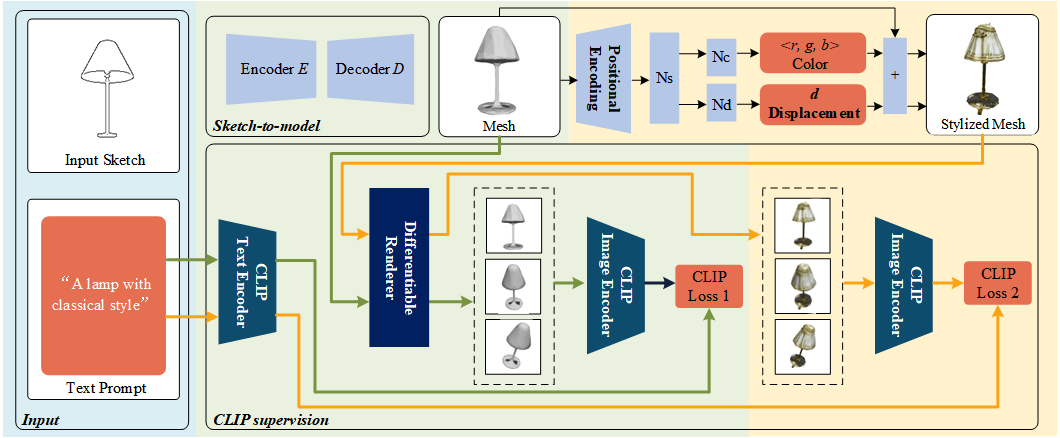}
	\caption{The Pipeline of Our Magic3DSketch. The Magic3DSketch takes a single-view sketch and text prompts to produce high-quality 3D objects. The neural network components in green background are designed to be generalized and trained end-to-end. The components in orange background are optimized on a per-object basis. }
	\label{fig:1}
\end{figure*}

As such, there is a need for a more accessible and intuitive approach to 3D modeling that can enable anyone to easily create high-quality 3D models without requiring extensive training or expertise. In this work, we explore the possibility of sketch-based 3D modeling, in which a novel 3D model is generated based on free-hand sketches from users. Although some recent works have enabled 3D modeling using text-based input alone \cite{ramesh2021zero,lin2022magic3d}, novice users may struggle to accurately describe their intended 3D models in detail, especially when it comes to complex or abstract shapes. Sketches, on the other hand, contain important spatial relationships and contextual information that convey the shape, size, and position of objects. Therefore, we believe sketch-based 3D modeling can be a promising solution that not only reduces the learning curve of 3D modeling but also provides greater creative freedom to accurately express  the creators' ideas and intentions.  

Despite the potential benefits of sketch-based 3D modeling, there are still significant challenges that need to be addressed. Sketches are inherently sparse and ambiguous, lacking color information and often representing only a simplified version of the intended object with boundaries. \textcolor{black}{Prior approaches incorporate meticulous line depictions from assorted angles or implement a gradual procedure reliant on \textit{strategic knowledge} \cite{cohen1999interface, deng2020interactive}, yet they prove to be challenging for beginners and tend to be laborious.} While there have been some recent developments in utilizing single-view sketches and auto-encoder deep neural networks to ease the 3D modeling process \cite{zhang2021sketch2model,guillard2021sketch2mesh}, these methods still face significant challenges in obtaining high-fidelity 3D models due to the inherent abstract and ambiguous nature of sketches, as well as the significant domain gaps between sketches and 3D shapes. Moreover, the lack of sufficient data for sketch-3D model pairs for large-scale training poses an additional challenge. As a result, existing deep learning methods for sketch-based 3D modeling are not be able to achieve satisfactory results. There is a pressing need for new approaches that can overcome these challenges.

To enhance the sketch-to-3D modeling process and address issues such as sparsity, ambiguity, and lack of training data, we propose to incorporate more \textit{prior knowledge} to the task. The emerging trend of Artificial Intelligence Generated Content (AIGC) and large-pretrained models such as GPT 
 \cite{brown2020gpt}, DALL-E \cite{betker2023dalle}, and Segment Anything (SAM) \cite{kirillov2023segment} has shed light to many AI tasks in language and vision \cite{bommasani2021opportunities}. Despite no popular large-scale models being trained explicitly for 3D tasks like our sketch-to-3D modeling, it is worth trying to leverage the knowledge gained from these models pre-trained on images and text to enhance the performance of our sketch-based 3D modeling process.

We hereby propose \textbf{Magic3DSketch}, a deep learning framework that generates high-fidelity 3D mesh from a single free-hand sketch with extra supervision from Image-Language Pre-Training. {\color{black} Along with the input sketch, a text prompt that describes the intended object is also inputted to incorporate some priors to the abstract and ambugious sketch. }The text prompt can be manually provided or automatically generated. In addition to the mesh silhouette constraint {\color{black} that forces the generated mesh to match the input sketch}, \textbf{Magic3DSketch} incorporates a Contrastive Language-Image Pre-training (CLIP) model to provide additional supervision with text-prompt and mesh rendering to the sketch-to-3D modeling process. Our pioneering endeavor to integrate text input and sketch input also has significant benefits for computer-human interaction. While text alone may not effectively convey detailed structural information, incorporating sketch input allows for better control over the structural attributes of the generated mesh, ensuring a closer match to users' desires. The single-view sketches can be sparse and ambiguous, necessitating additional information from texts to guide the generation of coherent shapes.

Our experiments have shown that incorporating Language-Image Pre-training can effectively enhance the performance of sketch-based 3D modeling. Our proposed \textbf{Magic3DSketch} has been tested on both synthetic and real hand-drawn datasets and has achieved state-of-the-art (SOTA) performance in 3D model generation in real-time. In contrast, conventional text-to-3D techniques typically take 30 minutes or longer to generate an optimized shape for each object \cite{lin2022magic3d}. Furthermore, our user study revealed that our sketch-based approach provides superior controllability and utility compared to utilizing text input alone with conventional text-to-shape methods. In the user study, users are also more satisfied with the 3D models generated by our approach than with existing approaches.

Furthermore, leveraging the CLIP model \cite{radford2021learning}, which learns a joint embedding space by mapping images and their corresponding textual descriptions to similar points in the embedding, we can now introduce colors to the sketch-based 3D modeling process. This allows our users to create more realistic and immersive models, enhancing visual aesthetics, and making them suitable for use in virtual and augmented reality, gaming, product design, and 3D digital asset creation.

In summarize, our contributions are as follows:
\begin{itemize}
\item We provide a novel way of 3D content creation by leveraging both text and sketch input. The approach enhances computer-human interaction by utilizing the strengths of each modality to generate coherent shapes that align with user preferences, compensating for the limitations of text in conveying detailed structural information and addressing the sparsity and ambiguity of single-view sketches. To the best of our knowledge, {\color{black}Language-Image Pre-training and text prompting are first used for the task of sketch-based 3D modeling. }
\item We have also demonstrated the capability of our proposed  \textbf{Magic3DS-} \textbf{ketch} with CLIP model in synthesizing textures for the generated mesh, which enables the creation of colored 3D models from a single freehand sketch and text prompt input of style or color description. Moreover, we have found that the CLIP-involved optimization in 3D model generation also enhances the model colorization process. This is the first time, to the best of our knowledge, that sketch-based 3D modeling with added colors to the model has been made possible.

\item Our experiment has demonstrated that we have achieved state-of-the-art (SOTA) performance in both synthetic and real datasets. Additionally, the sketch-to-model generation process can be performed in real-time ($>$ 100 FPS), which is significantly faster compared to existing text-to-3D approaches. Compared to existing text-to-3D approaches, our method provides users with increased controllability and utility, making it a more advantageous option. Moreover, in our user study, participants expressed greater satisfaction with the models generated by our \textbf{Magic3DSketch} approach compared to other methods.
\end{itemize}

\section{Related works}
\label{sec2}
\subsection{Sketch-Based 3D Modeling}

Sketch-based 3D modeling has been a topic of interest for researchers for decades, with significant advances being made in recent years \cite{bonnici2019sketch,olsen2009sketch}. There are two broad categories of existing sketch-based 3D modeling approaches: interactive and end-to-end. Interactive approaches require users to decompose the 3D modeling process into sequential steps or to use specific drawing gestures or annotations \cite{li2020sketch2cad,cohen1999interface,shtof2013geosemantic,deng2020interactive}. These approaches require strategic knowledge and may be challenging for users who are not familiar with the 3D modeling process. \textcolor{black}{Conversely, methods that employ an end-to-end strategy by utilizing template primitives or techniques based on retrieval yield satisfactory outcomes yet fall short in terms of adaptability} \cite{chen2003visual,wang2015sketch,sangkloy2016sketchy,8960471,10155453, nie2020m}.

In recent years, some researchers have approached the sketch-based 3D modeling problem as a single-view 3D reconstruction task, reconstructing the 3D model directly using deep learning \cite{zhang2021sketch2model,guillard2021sketch2mesh,chen2023deep3dsketch+,chen2023reality3dsketch,zang2023deep3dsketch+}. \textcolor{black}{Nonetheless, the disparity between sketch-based modeling and traditional single-view 3D reconstruction is pronounced. Given sketches' abstract and scant characteristics, coupled with an absence of textures, there's a necessity for extra hints to facilitate the creation of superior 3D forms.}

In this work, we tackle the challenge of lack of information by introducing models pre-trained at large scale datasets to provide additional supervision of producing realistic shapes. We also note that no prior works have been able to produce colored 3D model in sketch-based 3D modeling pipeline. Benefiting from pre-trained CLIP model that has learned texture information, we first introduce texture to the sketch-based 3D modeling. Sketch-based 3D modeling with colors is now made possible. 

\subsection{Single-View Reconstruction}

Following the introduction of extensive datasets such as ShapeNet\cite{chang2015shapenet}, methods driven by data have gained traction within the domain. Numerous data-driven strategies employ information at the category level to deduce 3D representations from single images\cite{chen2019learning,park2019deepsdf}. Additionally, some approaches seamlessly convert 2D images into 3D models\cite{liu2019soft,liu2019learning,kato2018neural,9817616}, significantly influenced by the advent of differentiable rendering technologies. Furthermore, the field has witnessed notable progress in unsupervised techniques for implicit function representation, largely attributed to advancements in differentiable rendering\cite{lin2020sdf,yu2021pixelnerf}.

Current approaches primarily concentrate on deriving 3D geometry from 2D images, our method is geared towards creating 3D meshes from 2D sketches, which are more abstract and less dense compared to real-world colored images. The production of high-quality 3D forms from such a simplified image representation remains an unresolved challenge. Unlike colored images, sketches lack detailed texture information and rely on the viewer's interpretation. This makes it difficult to create realistic 3D models that accurately reflect the original intent of the sketch. Additionally, sketches may not provide enough information for 3D reconstruction, as they are often incomplete or ambiguous. This work expands the application of existing single-view reconstruction framework to sketch-based 3D modeling and tackles the issue through text and language-image pre-trianing.

\subsection{3D Model Generation with CLIP}
The exploration of 3D model generation through the application of CLIP marks a significant advancement in the domains of computer vision and machine learning. The CLIP model, as delineated by Radford et al. \cite{radford2021learning}, is renowned for its proficiency in learning a joint embedding space that intricately intertwines image and text representations. This innovative model has revealed profound insights into the three-dimensional world, catalyzing a widespread upsurge in research that leverages 2D prior models to facilitate tasks in 3D generation. Forefront projects such as DreamField \cite{jain2022zero} and CLIP-Mesh \cite{mohammad2022clip} are emblematic of this trend, adopting a strategy of per-shape optimization. These endeavors predominantly focus on honing 3D representations, be it through neural radiance fields (NeRFs), meshes, or other forms, utilizing differentiable rendering to create 2D images from varied perspectives. Subsequently, these images are analyzed via the CLIP model to deduce loss functions \cite{jain2022zero, xu2023dream3d, jetchev2021clipmatrix, canfes2023text, lee2022understanding, hong2022avatarclip, mohammad2022clip}, which are instrumental in guiding the optimization of 3D shapes.The paper "Sketch2Model: View-Aware 3D Modeling from Single Free-Hand Sketches"\cite{zhang2021sketch2model} utilizes a view-aware generation approach within an extended encoder-decoder framework that includes a view auto-encoder, focusing on generating 3D meshes from single free-hand sketches with enhanced viewpoint conditioning.Additionally, the paper "Deep3DSketch+: Rapid 3D Modeling from Single Free-hand Sketches"\cite{chen2023deep3dsketch+} introduces an innovative end-to-end neural network approach for 3D modeling using only a single free-hand sketch. The method features a lightweight generation network for real-time inference and a structural-aware adversarial training strategy, which includes a Stroke Enhancement Module (SEM) to capture structural information and facilitate the learning of realistic and detailed shape structures.  Moving beyond mere shape optimization, various studies have embarked on training 3D generative models, capitalizing on the expansive embedding space offered by CLIP \cite{sanghi2022clip, liu2023iss++}, while others have focused their efforts on generating textures or materials for input meshes, utilizing the priors established by 2D models \cite{chen2023text2tex, michel2022text2mesh, wei2023taps3d}.

In our work, we harmonize the priors of CLIP with a differentiable renderer, aligning text prompts with 3D shapes within the collaborative vision-language embedding space of CLIP. This methodology steers our approach in the generation and colorization of 3D models, forging a path for more intricately nuanced and aligned 3D representations that are in harmony with their textual counterparts. The fusion of CLIP's robust capabilities with advanced 3D modeling techniques not only augments the fidelity of the generated models but also broadens their range of applications, spanning from artistic expression to pragmatic design implementations.

\section{Method}
\label{sec3}
In this section, we first introduce the encoder-decoder backbone of our \textbf{Magic3DSketch} (Sec. \ref{sec:backbone}). We then introduce the integration of CLIP model and text prompt for extra supervision (Sec. \ref{sec:supervision}). Eventually, we demonstrate the approach to colorize or stylize the generated mesh with text prompts and CLIP supervision (Sec. \ref{sec:style}).
\subsection{Preliminary}
For the sketch-based 3D modeling task, in this work, we use a single binary sketch ${I\in \left\{0,1\right\}^{W\times H}}$ as the input of 3D modeling. We let ${I \left [ i,j \right ] = 0 }$ if marked by the stroke, and ${I\left [ i,j \right ] = 1 }$ otherwise. The network $G$ is designed to obtain a mesh ${M_\Theta =(V_\Theta,  F_\Theta)}$, in which ${V_\Theta}$ and ${F_\Theta}$ represents the mesh vertices and facets, and the silhouette ${S_\Theta :\mathbb{R}^3 \rightarrow \mathbb{R}^2} $ of ${M_\Theta}$ matches with the input sketch $I$.


\subsection{The Encoder-Decoder Backbone}\label{sec:backbone}
As illustrated in Fig. \ref{fig:1}, the backbone of our \textbf{Magic3DSketch} is an encoder-decoder structure to transform sketch to a 3D mesh. Specifically, the encoder $E$ is responsible for transforming the input sketch into a latent shape code $z_s$, which summarizes the sketch on a coarse level by considering its semantic category and conceptual shape. The decoder $D$ then takes the latent shape code $z_s$ and uses it to generate the mesh $M_\Theta = D(z_s)$. Unlike some previous methods {\color{black} \cite{xu2019disn, chen2024deep3dsketchim}} that use structures like MLPs to predict point-wise locations, our method employs cascaded upsampling blocks to calculate the vertex offsets of a template mesh, which is then deformed to obtain the output mesh $M_\Theta$. This approach enables our method to gradually infer 3D shape information with increased spatial resolution to obtain more details of the output mesh $M_\Theta$.

To supervise the generation process, we render the generated mesh $M_\Theta$ with a differentiable renderer to produce a image of silhouette $S_\Theta$. This allows us to train the network end-to-end using the supervision of the rendered silhouettes, with approximated gradients of the differentiable renderer. The rendered silhouettes are compared with input sketches with the metric IoU calculated $\mathcal{L}_{iou}$, which is defined as:
\begin{align}
\mathcal{L}_{i o u}\left(S_{1}, S_{2}\right)=1-\frac{\left\|S_{1} \otimes S_{2}\right\|_{1}}{\left\|S_{1} \oplus S_{2}-S_{1} \otimes S_{2}\right\|_{1}}
\end{align}
where $S_1$ and $S_2$ is the rendered silhouette. 

To improve the computational efficiency, we progressively increase the resolutions of silhouettes and derives the multi-scale mIoU loss {\color{black} $\mathcal{L}_{ms}$}, which is represented as: {\color{black} 
\begin{align}
\mathcal{L}_{ms}=\sum_{i=1}^{N} \lambda_{s_{i}} \mathcal{L}_{iou}^{i}
\label{lsp}
\end{align}
}

As demonstrated in previous sketch-to-model works \cite{zhang2021sketch2model,guillard2021sketch2mesh}, using the encoder-decoder structure allows the neural network to generate a coarse shapes from sketches. In the next section, we introduce how we integrate CLIP and text prompt for extra supervision to obtain more detailed models with higher fidelity.

\subsection{Sketch-to-Mesh with CLIP Discriminator} \label{sec:supervision}
We then introduce the ``magic" brought by CLIP through a CLIP Discriminator. Specifically, we guide our neural optimization using the joint vision language embedding space of CLIP. A text prompt $T$ describing the desired generated object is inputted to our framework. We embed text prompt into CLIP space by letting the text prompt $T$ go though a CLIP Language Encoder $E_{L1}$ to form a latent code $E_{L1}(T)$. 

Meanwhile, the image obtained from differentiable renderer $S_\Theta$ is also embedded into CLIP space by passing another CLIP Image Encoder $E_{I}$ to form a latent code $E_{I}(S_\Theta)$. The CLIP discriminator's goal is to force the generated mesh to match the text prompt. Therefore, we calculate the CLIP Loss 1 $\mathcal{L}_{CLIP}$ as:
\begin{align}
\mathcal{L}_{CLIP}\left(S_\Theta, T_{1}\right)=1-\left \langle E_{I}(S_\Theta), E_{L1}(T) \right \rangle 
\end{align}
where $\left \langle  \right \rangle $ is the  cosine similarity operator.
To fully utilize the capability of CLIP, we propose to utilize multi-view supervision to enhance its understanding of the generated mesh and improve overall robustness. As many previous works investigated in the realm of shape-from-silhouette, the proposed multi-view silhouettes contain valuable geometric information about the 3D object~\cite{gadelha2019shape,hu2018structure}. We randomly sample $N$ camera poses $\xi_{1...N}$ from camera pose distribution $ p_{\xi} $. In each training iteration, the CLIP encoder sees the object from different views. We use the differentiable renderer to render the silhouettes $S_{1...N}$ from the mesh $M$ and render the silhouettes $S_r\left \{1...N \right \} $ from the mesh $M_r$. Next, we compute the average of its encoding from multiple augmented views to form the $\mathcal{L}_{CLIP1}$.

\subsection{Viewpoint Prediction Module}
In our approach, we utilize an encoder network denoted as $E$ to generate a latent code $z_l$, which is then fed into the viewpoint prediction module. The viewpoint prediction module consists of two fully-connected layers represented by $D_v$, which predict the viewpoint estimation $\xi_{pred}$ in the form of Euler angles. This viewpoint prediction module is trained in a fully-supervised manner, with the ground truth viewpoint $\xi_{gt}$ as input and supervised by a viewpoint prediction loss $\mathcal{L}_{v}$. The loss function, in this case, adopts the Mean Squared Error (MSE) loss to measure the discrepancy between the predicted and ground truth poses, which is defined as:
\begin{align}
\mathcal{L}_{v}=\|\xi_{gt}-\xi_{pred}\|_{2}=\left\|\xi_{gt}-D_{v}\left(z_{l}\right)\right\|_{2}
\end{align}

\subsection{Loss Function Components and Weights}\label{sec:loss}

The loss function $\mathcal{L}$ for the sketch-to-model process is calculated as the weighted sum of four components:
\begin{align}
\mathcal{L} = {\color{black} \lambda_{ms} \mathcal{L}_{ms}} + \lambda_{r} \mathcal{L}_{r} + \lambda_{clip} \mathcal{L}_{CLIP} + \lambda_{v}\mathcal{L}_{v}
\label{loss}
\end{align}
$\mathcal{L}_{r}$ denotes the flatten loss and Laplacian smooth loss as in~\cite{zhang2021sketch2model, kato2018neural, liu2019soft}, which is used to make meshes more realistic with higher visual quality. The weights $\lambda_{r}$, $\lambda_{v}$, {\color{black} $\lambda_{ms}$}, and $\lambda_{clip}$ correspond to the four components, respectively.
\subsection{Stylization for Colored Mesh with CLIP}\label{sec:style}
The incorporation of color information can be seen as an additional layer of information to the 3D model. Consequently, a two-stage approach is employed for both 3D model generation and colorization. The stylization or colorization is performed after the 3D model has been obtained. However, we find that the CLIP enabled optimization in the shape generation process can be beneficial to the later stylization or colorization stage. 

The 3D model stylization is also performed with CLIP supervision. Here, we follow the work \cite{michel2022text2mesh} that not only edit the color of the 3D model, but also edit the shape of it according to user input styles. We assign a style attribute for every point in a mesh represented by neural fields. For every point $p \in V$, we apply a positional encoding using Fourier feature mappings to obtain the neural field represented by an MLP $N_s$. The MLP $N_s$ output is a color and displacement $(c_p, d_p) \in (\mathbb{R^3}, \mathbb{R})$ along the surface normal. The $N_s$ branches out to MLPs $N_d$ and $N_c$, representing the a displacement along the vertex normal and the RGB color information. The stylized mesh prediction $M_{style}$ has every point $p$ displaced by the output of $N_d$ and colored by the output of $N_c$. Vertex colors propagate over the entire mesh surface using an interpolation-based differentiable renderer to obtain the rendered images $S_{style}$ for optimization. 

The goal of the optimization is the $S_{style}$ matches the input text prompt. This process is also enabled by CLIP and is optimized at CLIP space. Specifically, the $S_{style}$ goes through the CLIP image encoder and obtain the image feature in CLIP space. The image feature is compared with the text feature in CLIP space derived from the text-prompt $T_{2}$ by CLIP Language Encoder. A cosine similarity of CLIP text feature and CLIP image feature is calculated and is used for the loss function $\mathcal{L}_{Style}$ of stylization. 
\begin{align}
\mathcal{L}_{Style}\left(S_{style}, T_{2}\right)=1-\left \langle E_{I}(S_{style}), E_{L}(T_{2}) \right \rangle 
\end{align}
As illustrated in the Fig. \ref{fig:1}, similar multi-view augmentation is performed and the final loss function is the average of the CLIP loss across all viewpoints. 
\begin{table}[!h]
\begin{center}
		\caption{The Voxel IoU of ShapeNet-Synthetic and ShapeNet-Sketch dataset}
   \label{table.Syntheticiou}
\resizebox{\textwidth}{!}{
\begin{tabular}{cllllllllllllll}
\hline
\multicolumn{15}{c}{ShapeNet-Synthetic / ShapeNet-Sketch (Voxel IoU$\uparrow$)}                                                                                                                                                                                                                                                                  \\
\multicolumn{1}{l}{} & \multicolumn{2}{c}{car}                    & \multicolumn{2}{c}{sofa}                   & \multicolumn{2}{c}{airplane}               & \multicolumn{2}{c}{bench}         & \multicolumn{2}{c}{display}                         & \multicolumn{2}{c}{chair}                  & \multicolumn{2}{c}{table}         \\ \hline
Retrival             & 0.667                     & 0.626          & 0.483                     & 0.431          & 0.513                     & 0.411          & 0.380                     & 0.219 & 0.385                              & 0.338          & 0.346                     & 0.238          & 0.311                     & 0.232 \\
Auto-Encoder         & 0.769                     & 0.648          & 0.613                     & \textbf{0.534} & 0.576                     & 0.469          & 0.467                     & 0.347 & 0.541                              & 0.472          & 0.496                     & 0.361          & \textbf{0.512}            & 0.359 \\
Sketch2Model         & \multicolumn{1}{c}{0.751} & 0.659          & \multicolumn{1}{c}{0.622} & \textbf{0.534} & \multicolumn{1}{c}{0.624} & 0.487          & \multicolumn{1}{c}{0.481} & 0.366 & \multicolumn{1}{c}{\textbf{0.604}} & 0.479 & \multicolumn{1}{c}{0.522} & 0.393          & \multicolumn{1}{c}{0.478} & 0.357 \\
Deep3DSkitch+        & 0.728                     & 0.675          & \textbf{0.640}            & 0.534          & 0.632                     & 0.490          & \textbf{0.510}            & 0.368 & 0.588                              & 0.463          & 0.525                     & 0.382          & 0.510                     & 0.370 \\

\textbf{Ours}        & \textbf{0.772}            & \textbf{0.688} & 0.639                     & 0.531          & \textbf{0.637}            & \textbf{0.497} & 0.477                     & 0.382 & 0.593                              & \textbf{0.497}       & \textbf{0.530}            & \textbf{0.413} & 0.503                     & \textbf{0.372}
\end{tabular}}
\resizebox{\textwidth}{!}{
\begin{tabular}{cllllllllllllll}
\hline
\multicolumn{1}{l}{} & \multicolumn{2}{c}{telephone}              & \multicolumn{2}{c}{cabinet}                         & \multicolumn{2}{c}{loudspeaker} & \multicolumn{2}{c}{watercraft}  & \multicolumn{2}{c}{lamp}        & \multicolumn{2}{c}{rifle}       & \multicolumn{2}{c}{mean}        \\ \hline
Retrival             & 0.667                     & 0.536          & 0.518                              & 0.431          & 0.468          & 0.365          & 0.422          & 0.369          & 0.325          & 0.223          & 0.475          & 0.413          & 0.455          & 0.370          \\
Auto-Encoder         & 0.706                     & 0.537          & 0.663                              & 0.534          & 0.629          & 0.533          & 0.556          & 0.456          & 0.431          & 0.328          & 0.605          & 0.541          & 0.582          & 0.372          \\
Sketch2Model         & \multicolumn{1}{c}{0.719} & 0.554          & \multicolumn{1}{c}{\textbf{0.701}} & \textbf{0.568} & \textbf{0.641} & \textbf{0.544} & \textbf{0.586} & 0.466 & \textbf{0.472} & 0.338          & 0.612          & 0.534          & 0.601          & 0.483          \\
Deep3DSkitch+        & \textbf{0.757}            & \textbf{0.576} & 0.699                              & 0.553          & 0.630          & 0.514          & 0.583          & \textbf{0.467}          & 0.466          & \textbf{0.347} & \textbf{0.632}          & 0.543          & \textbf{0.611} & 0.483          \\

\textbf{Ours}        & 0.697                     & 0.517          & \textbf{0.701}                     & 0.555          & 0.620          & 0.523          & 0.572          & 0.459          & 0.459          & 0.328          & \textbf{0.632} & \textbf{0.550} & 0.602          & \textbf{0.484} \\ \hline
\end{tabular}}
\end{center}
\end{table}

\section{Experiments and results}
\label{sec4}

\subsection{Implementation Details}
Our \textbf{Magic3DSketch} pipeline consists of two separate processes, as shown in Fig. \ref{fig:1}. The first process is the sketch-to-model generation, which is trained end-to-end to rapidly generate 3D models from sketches. The second process is the shape stylization, which is performed on a per-object basis for optimization. By separating these processes, we can achieve both speed and flexibility in generating high-quality 3D models with various styles. We utilize ResNet-18 \cite{he2016deep} as the encoder for image feature extraction. The extracted 512-dim feature is processed through two linear layers with L2-normalization, yielding a 512-dim shape code zs and a 512-dim view code zv. The template mesh of the encoder-decoder structure is a sphere. The rendering module is SoftRas \cite{liu2019learning}, and the number of views N = 3 under a uniform camera distribution. Each 3D object is positioned in the canonical view with a set distance from the camera and 0 in evaluation and 0 in azimuth angle. The text prompt T for every [category] is the same, which is “A grey [category]”. We utilize the Adam optimizer with an initial learning rate of 1e-4 and multiply by 0.3 every 800 epochs. Betas are set to 0.9 to 0.999. $\lambda_{r}$,  $\lambda_{v}$, $\lambda_{sp}$ and $\lambda_{clip}$ in Equation \ref{loss} equal to 0.1. The total number of training epochs is 2000.
\subsection{Experimental Result for Sketch-to-Mesh Generation}

We conducted an evaluation of our sketch-to-mesh by comparing the performance of our method with several other methods, including a naive autoencoder network, model retrieval with features from a pre-trained sketch classification network, and the Sketch2Model \cite{zhang2021sketch2model} as the existing state-of-the-art (SOTA) model. We used the ShapeNet-Synthetic dataset, which provided accurate ground truth 3D models for evaluation purposes. ShapeNet-Synthetic consists of edge maps extracted by a Canny edge detector from rendered images provided by Kar et al. \cite{kar2017learning}, encompassing thirteen distinct types of 3D models from ShapeNet. Additionally, we tested the performance on the ShapeNet-Sketch dataset using the same evaluation method to validate the model’s performance on real human drawings. The ShapeNet-Sketch dataset comprises 1300 sketches and their corresponding 3D models, drawn by participants based on 3D object renderings from Kar et al.'s dataset. The results for the 3D reconstruction metrics, voxel IoU, are detailed in Table \ref{table.Syntheticiou}. Our method outperforms existing methods and achieves state-of-the-art (SOTA) performance.

Furthermore, during inference, camera pose data is not necessarily required since the camera poses can be learned from synthetic data and applied to real-world sketches. To validate this, we tested our approach on the ShapeNet-Sketch dataset, with the results presented in Table \ref{table: pose}. The elevation and azimuth data are the poses predicted by the network for the sketches. Additionally, our method utilizes both predicted and ground truth viewpoints to generate shapes, which we then evaluated using voxel IoU and compared against Sketch2Model. The results are displayed in Table \ref{table:compare}. "GT Pos" indicates the use of ground truth viewpoints as input, while "Pred Pos" denotes the use of predicted viewpoints. The average voxel IoU for shapes generated using predicted viewpoints by our method is 0.599, compared to 0.602 when using real viewpoints, showing a minimal difference of 0.03. This demonstrates the superior viewpoint estimation capability of our method.
		
 
 

\begin{table*}[ht]
\begin{center}
\caption{{The quantitative evaluation of sketch-view estimation}}
\label{table: pose}
\resizebox{\textwidth}{!}{
\begin{tabular}{c c c c c c c c c}
\hline
\multicolumn{9}{c}{Shapenet-Sketch (MAE $\downarrow)$} \\

\multicolumn{2}{c}{}  & car & sofa & airplane & bench & display & chair & table \\
 \hline
\multirow{2}{*}{Elevation} & Sketch2Model & 1.0751 & 1.5989 & 2.3899 & 1.8345 & 1.8944 & 1.8690 & 1.2857  \\
 & \textbf{Ours} & \textbf{0.9029} & \textbf{1.395} & \textbf{2.2014} & \textbf{1.0168} & \textbf{1.6826} & \textbf{1.5422} & \textbf{1.0184} \\
 
\multirow{2}{*}{Azimuth} & Sketch2Model & 5.0986 & 11.0327 & 10.4171 & 43.7923 & 44.1861 & 8.6753 & \textbf{86.7654}  \\
 & \textbf{Ours} & \textbf{4.3056} & \textbf{9.8532} & \textbf{9.7180} & \textbf{38.7755} & \textbf{43.2417} & \textbf{7.2630} & 87.6369 
 \\
 
\hline
\multicolumn{2}{c}{} & telephone & cabinet & loudspeaker & watercraft & lamp & rifile & mean \\
\hline
\multirow{2}{*}{Elevation} & Sketch2Model & 2.2732 & 1.2148 & 2.4303 & 3.6884 & 4.4071 & 3.3226 & 2.2526 \\
 & \textbf{Ours} & \textbf{2.072} & \textbf{1.0168} & \textbf{2.0659} & \textbf{3.4014} & \textbf{4.0796} & \textbf{3.1199} & \textbf{1.9627} \\

\multirow{2}{*}{Azimuth} & Sketch2Model & 54.3659 & 41.7126 & 73.8672 & 34.5512 & \textbf{84.4146} & 11.2999 & 39.2445  \\
 & \textbf{Ours} &  \textbf{51.2568} & \textbf{38.7755} & \textbf{72.6029} & \textbf{33.8180} & 84.7734 & \textbf{10.7342} & \textbf{37.9042} \\
\hline
\end{tabular}}
\end{center}

\end{table*}
Moreover, with predicted viewpoints, our method attained a mean voxel IoU score of 0.599, outperforming Sketch2Model’s score of 0.584. This validates the superior performance of our approach on real-world sketches without requiring known viewpoints.

\begin{table}[!h]

	\begin{center}
		\caption{Performance Comparison of Shape Generation Using Predicted and Ground Truth Viewpoints on ShapeNet-Sketch Dataset}
   \label{table:compare}

\scalebox{0.7}{
\begin{tabular}{c c c c c c c c c}
\hline
\multicolumn{9}{c}{Shapenet-synthetic (Voxel IoU $\uparrow)$} \\

\multicolumn{2}{c}{} & car & sofa & airplane & bench & display & chair & table \\
\hline
\multirow{2}{*}{} 
&Sketch2Model(GT Pos) & 0.751 & 0.622 & 0.624 & \textbf{0.481} & \textbf{0.604} & 0.522 & 0.478\\
 & Sketch2Model(Pred Pos) & 0.746 & 0.620 & 0.618 & 0.477 & 0.550  & 0.515 & 0.470  \\
 & \textbf{Ours (GT Pos)} & \textbf{0.772} & \textbf{0.639} & \textbf{0.637} & 0.477 & 0.593 & \textbf{0.530} & \textbf{0.503} \\
 & \textbf{Ours (Pred Pos)} & 0.771 & 0.638 & 0.634 & 0.476 & 0.579 & 0.528 & 0.502  \\
\hline
\multicolumn{2}{c}{} & telephone & cabinet & loudspeaker & watercraft & lamp & rifle & mean \\
\hline
\multirow{2}{*}{} 
& Sketch2Model(GT Pos) &  \textbf{0.719} & \textbf{0.701} & \textbf{0.641} & \textbf{0.586} & \textbf{0.472} & 0.612 &0.601\\
& Sketch2Model(Pred Pos) & 0.673 & 0.667 & 0.624 & 0.569  & 0.463 & 0.606 & 0.584  \\
& \textbf{Ours (GT Pos)} & 0.697   & \textbf{0.701} & 0.620     & 0.572 & 0.459 & \textbf{0.632} & \textbf{0.602} \\
& \textbf{Ours (Pred Pos)} & 0.693  & 0.700 & 0.619   & 0.570 & 0.456 & 0.631 & 0.599 \\
\hline
\end{tabular}}

  

    
	\end{center}

\end{table}
\vspace{-\baselineskip}

Moreover, the qualitative analysis of the generated models demonstrated the effectiveness of our approach in generating more realistic and high-fidelity structures. Fig. \ref{fig:lk} and Fig. \ref{fig:sketch} present a visual comparison between the models generated by our proposed method and the existing state-of-the-art (SOTA) approach on the ShapeNet-Synthetic and ShapeNet-Sketch datasets, respectively. These comparisons showcase the superior performance of our approach in terms of overall structure and shape fidelity. Specifically, our method successfully reconstructed detailed structures such as the thin legs of chairs and desks, as well as the thin lamp stand in the lamp, in accordance with the input sketches. Our approach is able to produce models with higher levels of geometric complexity, such as the spoilers of the car and the engine of the airplane. Moreover, the generated models from our approach also exhibit better consistency between different parts of the model, resulting in more coherent and visually pleasing structures overall. Additionally, we demonstrated impressive visual results on real-world sketches, indicating that our method, trained on synthetic data, can successfully generalize to real sketch data.
\begin{figure*}[!h]
	\centering
	\includegraphics[width=1\textwidth]{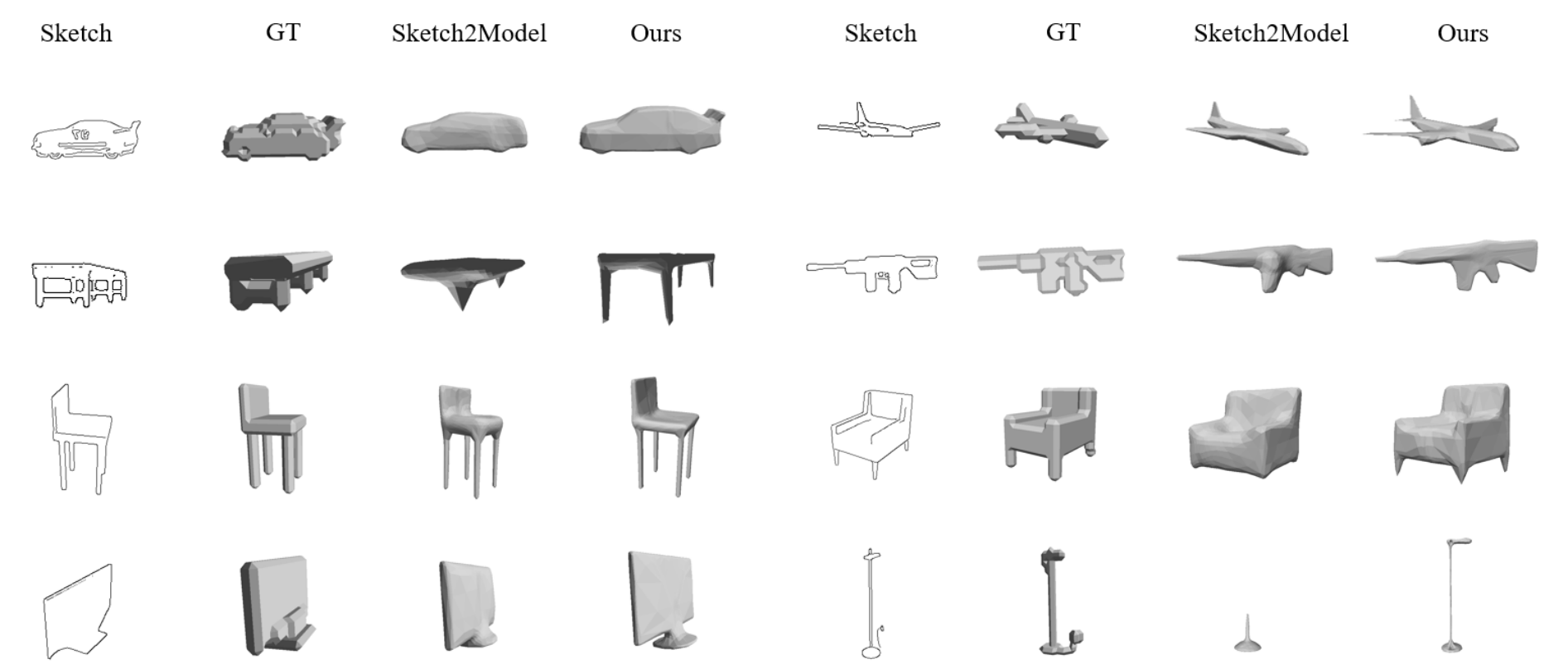}
	\caption{{\textbf{Qualitative evaluation with existing state-of-the-art sketch-to-model approaches.} The visualization of 3D models generated demonstrated that our method is capable of synthesizing higher fidelity and more realistic 3D models. }}
	\label{fig:lk}
\end{figure*}

\begin{figure}[!h]
\centering
\includegraphics[width=0.7\textwidth]{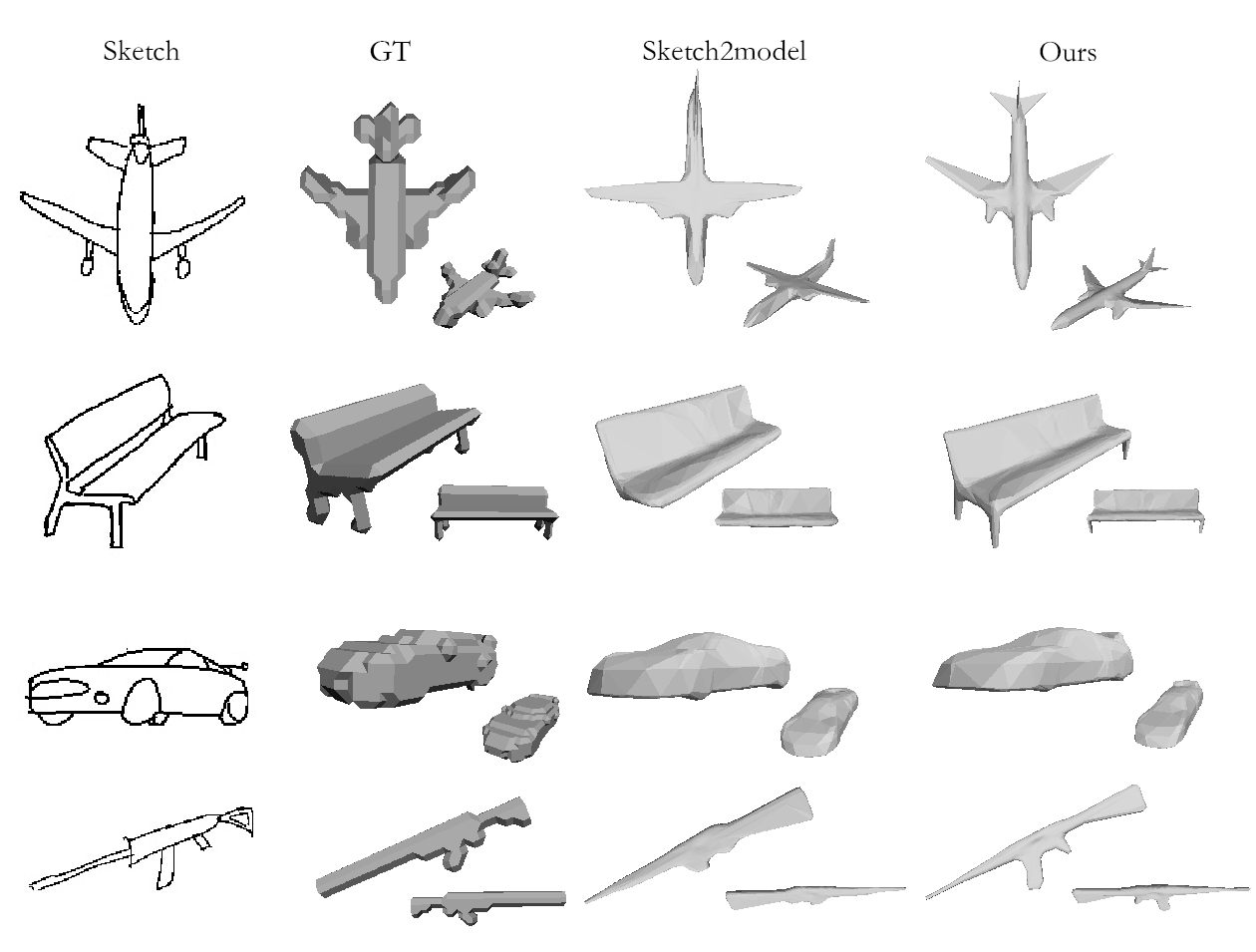}
\caption{\textbf{Representative results on our ShapeNet-Sketch dataset.} 
{\color{black}
In comparison to alternative methods, Magic3DSketch boasts the capability to create more advanced and promising shapes using hand-drawn sketches}}
\label{fig:sketch}
\end{figure}

\begin{figure*}[ht]
	\centering
	\includegraphics[width=1\textwidth]{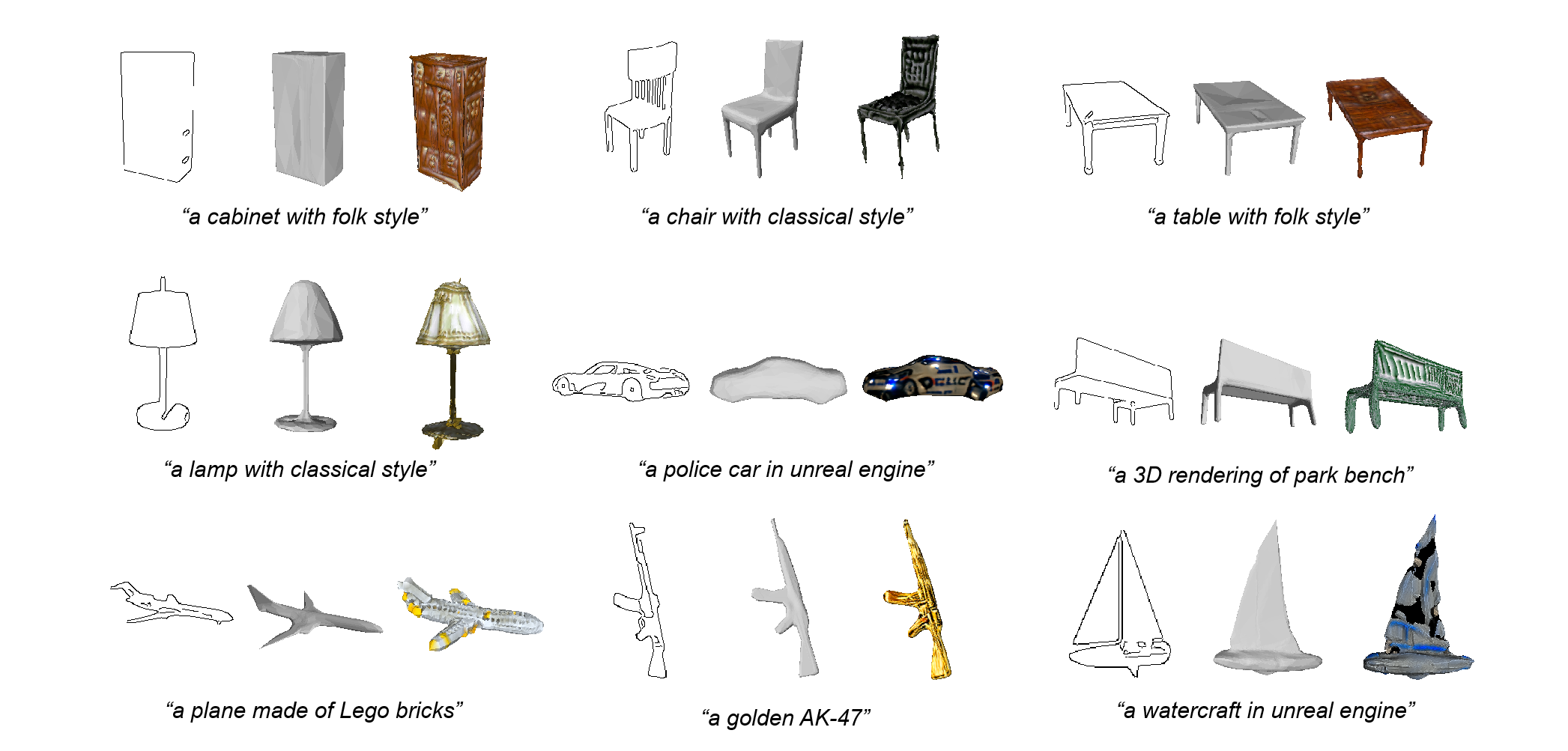}
	\caption{The visualization of 3D models generated demonstrated that our method is capable of synthesizing high quality 3D objects with shapes from sketches and text prompts}
	\label{fig:2}
\end{figure*}

\begin{figure}[!h]
\centering
\includegraphics[width=0.8\textwidth]{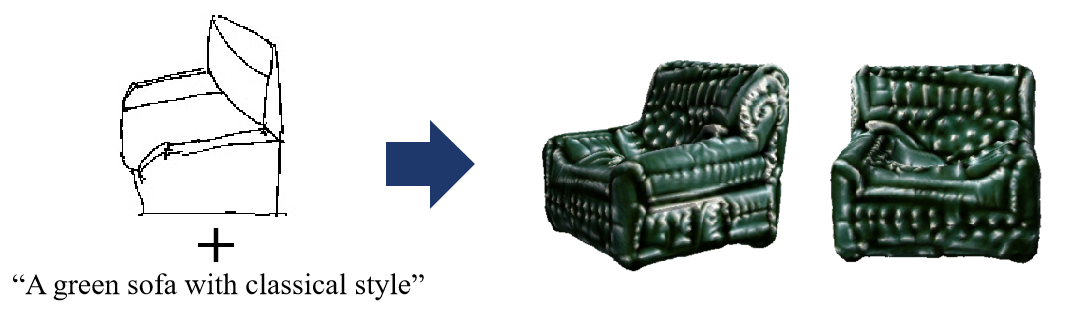}
\caption{
{\color{black}
Generation of High-Quality and colorized 3D Models Using Real Hand-Drawn Sketches and Text Prompts via Our Method}}
\label{fig:skco}
\end{figure}

\subsection{Stylization After Sketch-to-Mesh Conversion}
We evaluated the ability of our approach to generate stylized 3D models that match the intended style described in the text prompt while preserving the fidelity of the original 3D model. We provided different text prompts and sketches across various categories, with the results shown in Fig. \ref{fig:2}. We also tested our method using real hand-drawn sketches, as illustrated in Fig. \ref{fig:skco}. This demonstrates a use case where a user draws a free-hand sketch with a text prompt and obtains a textured 3D shape. The results indicate that the CLIP-enabled stylization successfully produces colored and realistic 3D objects.

\subsection{Runtime Evaluation}

The speed of the algorithm is always an issue in 3D modeling. Our sketch-to-model process in \textbf{Magic3DSketch} is a generalized solution, which is significantly faster than the NeRF based approaches that performs per-object optimization \cite{jain2022zero,lin2022magic3d} which usually takes tens of minutes or even multiple hours to obtain a single 3D model. Our method is even faster than other generalized sketch-based 3D modeling approaches. \textcolor{black}{Our method was subjected to a quantitative analysis to confirm our framework's efficacy, as illustrated in the accompanying Table \ref{table: Runtime} Following comprehensive training of our framework, the neural network underwent testing on a desktop equipped with a mainstream graphics processor (NVIDIA GTX 3090), achieving a rendering velocity of 129 Frames Per Second (FPS) when utilizing GPU acceleration. This demonstrates a notable enhancement in speed, exceeding Sketch2Model \cite{zhang2021sketch2model} by more than 43$\%$ under identical experimental conditions (0.018s). \begin{table}[ht]
	\begin{center}
  \caption{The Average Runtime for Sketch-to-Model Conversion}
  \label{table: Runtime}	
		\scalebox{0.8}{
			\setlength{\tabcolsep}{7mm}
			\begin{tabular}{ c  c  c  }
				\hline
				& Speed(s) & FPS \\ \hline
				Inference by GPU & 0.0078   & 129 \\ \hline
				Inference by CPU & 0.0205   & 49  \\ \hline
		\end{tabular}}
 
	\end{center}

\end{table}
Additionally, an evaluation exclusively on CPU (Intel Xeon Gold 6326 CPU @ 2.90GHZ) yielded a throughput of 49 FPS, indicating its adequacy for real-time interactions between computers and humans.}

{\color{black}\subsection{User Study for Qualitative Evaluation}}
Next, we conducted another user study to verify the image quality of our generated images. We conducted a user study following the settings of \cite{cai2021unified,michel2022text2mesh,yao2022dfa,zhang2023painting} and used the metric of widely-used Mean Option Score (MOS) ranging from 1-5 \cite{seufert2019fundamental} to the following two factors:
\begin{enumerate}
\item Q1:  How well does the output 3D model match the input sketch? \textit{(Fidelity)} ; 
	\item Q2:  How do you think the quality of the output 3D model? \textit{(Quality)}.

 \end{enumerate}

We recruit 12 experienced designers to participate in the study. Each participant was given a brief introduction to the concept of fidelity and quality before being presented with 36 generated 3D modeling results produced by our algorithm. We asked participants to rate each model on a scale of 1 to 5. 
The results of the user study are summarized in Table \ref{table: Opinion}, where we report the average scores across all participants. The results demonstrate that our method outperforms the existing state-of-the-art method in terms of users' subjective ratings. These findings provide additional evidence for the effectiveness of our approach in generating high-quality and realistic 3D models.
\begin{table}[!h]
	\setlength{\tabcolsep}{6mm}
	
	\begin{center}
   \caption{Mean Opinion Scores (1-5) with 5 being the highest rating for the perceived fidelity and quality of the model, or Q1 (Fidelity) and Q2 (Quality).}
  \label{table: Opinion}
		\scalebox{0.85}{
			\begin{tabular}{l c c}
				\hline
				& {(Q1): Fidelity} & {(Q2): Quality} \\ \hline
				
				Sketch2Model & 3.09                               & 3.01         \\                     
				Ours         & \textbf{3.81}                      & \textbf{3.22}                     \\ \hline
		\end{tabular}}

	\end{center}
	
\end{table}

{\color{black}\subsection{User Study for Content Creation}}
To validate the superiority of controllability, we also performed additional user experiments by recruiting 11 3D designers from a 3D printing company. We allowed the designers to input their desired shape information by text and run the baseline models \textbf{Magic3D} \cite{lin2022magic3d}. They also described and sketched their desired models and used our approach to generate textured 3D shapes. \textcolor{black}{Participants were requested to assess the controllability and usefulness of each method, these being key factors frequently considered in the assessment of user interface usability and user experience} \cite{albert2022measuring} \cite{oh2018lead}. Following the settings in \cite{albert2022measuring}, we used a 7-point Likert scale ranging from highly disagree to highly agree. The results are shown in Table \ref{table:Controllability}. Users give higher ratings for both controllability and usefulness scores with our method compared to existing text-to-3D approaches.

\begin{table}[h]
    \centering
    \caption{The qualitative evaluation of Controllability}
    \label{table:Controllability}
    \setlength{\tabcolsep}{6mm}
    \scalebox{0.85}{
        \begin{tabular}{lcc}
            \hline
            & (Q1): Controllability & (Q2): Usefulness \\
            \hline
            Magic3D & 2.45 $\pm$ 0.69 & 3.36 $\pm$ 0.81 \\
            Ours & \textbf{5.72 $\pm$ 0.65} & \textbf{5.82 $\pm$ 0.98} \\
            \hline
        \end{tabular}
    }
\end{table}

\subsection{Ablation Study for CLIP Supervision}
Unlike shape-from-silhouette or multi-view stereo methods that use multi-view information, we only use a single-view sketch as input. However, we have found that the silhouette alone cannot convey all the necessary information, as the IoU loss only enforces silhouette matching in a single view. This means that the single-view sketch/silhouette lacks 3D information, making many parts of the 3D model unobservable. To address this limitation, we apply CLIP supervision to multi-view silhouettes, which do not have a ground truth sketch of the user's intention. If we were to simply add ground truth silhouettes for supervision from the 3D model (synthesized silhouettes from other views), the model would have to learn additional "camera pose" information (the silhouette properties at different viewpoints), which is difficult to optimize and can result in suboptimal results. Instead, we found that CLIP supervision can effectively fill in (or imagine) the hidden parts of the model in a reasonable manner without the need to learn camera pose information, as the CLIP model has seen many objects in the real world from different viewpoints. As a result, the generated shapes are more plausible and reasonable, which is better aligned with the theme of our creative sketch-to-model task.

Firstly, we show that CLIP supervision is effective in sketch-to-model process. We conducted the ablation study that removes the CLIP supervision by removing the CLIP Loss. Our quantitative result (Table \ref{table: ablation iou}) shows removing the CLIP supervision be detrimental to the performance for both predicted viewpoint and ground-truth viewpoint settings. 

{\color{black}
	
To validate the effectiveness of our learning strategy, we calculated the average CLIP score for the "sofa" class with randomly sampled views (uniform sampling). The results (Table \ref{table:ablation cs}) show that our method can effectively learn shapes that are more plausible and reasonable, as represented by the higher CLIP score.} The results in Fig. \ref{fig:ablation} further illustrate the effectiveness of CLIP loss in obtaining fine structures and obtain realistic shapes. After removing the CLIP supervision, some details of the model is missing, such as the part of the boat. CLIP supervision is capable of producing models with higher levels of geometric complexity and make generated models exhibit better consistency between different parts of the model, resulting in more coherent and visually pleasing structures overall, as evident in the chair example.
\begin{figure*}[!h]
	\centering
	\includegraphics[width=1\textwidth]{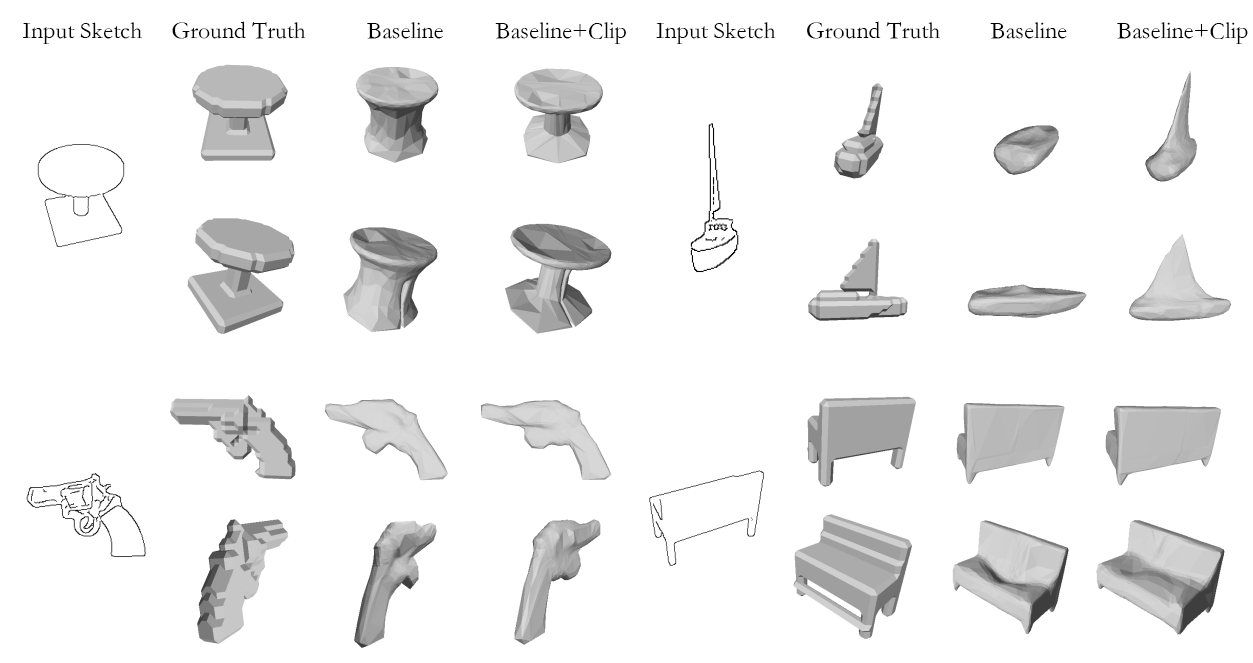}
	\caption{The visualization of Ablation Study. The absence of CLIP supervision in sketch-to-model process results in the loss of certain details in the 3D model, such as the sail of the boat, and the creation of unrealistic structures, such as the sitting pad of the chair model.}
	\label{fig:ablation}
\end{figure*}
\begin{table}[ht]
	\begin{center}
		\caption{The quantitative evaluation of ablation study of Shape Generation with CLIP supervision}
  \label{table: ablation iou}
			
		\scalebox{0.85}{
			\begin{tabular}{lc}
				\hline
				\multicolumn{2}{c}{Shapenet-synthetic (Voxel IoU $\uparrow)$} \\
				\hline
				w/o CLIP supervision (GT Pos)  & 0.6004\\
				w/o CLIP supervision (Pred Pos)  &  0.5978 \\
				w/ CLIP supervision (GT Pos)&  \textbf{0.6024} \\
				w/ CLIP supervision (Pred Pos)& 0.5999 \\
				\hline
		\end{tabular}}
  
	\end{center}
\end{table}
\begin{table}[ht]
	\begin{center}
		\caption{The quantitative evaluation of ablation study for CLIP Supervision for textured 3D models}
  \label{table:ablation cs}
		
		\scalebox{0.8}{
			\begin{tabular}{ccc}
				\hline
				~ & No CLIP Supervision
 & Use CLIP Supervision
 \\
				\hline
				CLIP Score & 0.2962 & \textbf{0.3044}\\
				\hline
		\end{tabular}}
  
	\end{center}
\end{table}

\subsection{Ablation Study For CLIP Supervision to Mesh Stylization} 

We contend that even though sketch-to-model and mesh stylization are two distinct processes, the CLIP supervision used in the former stage is also beneficial for the latter. To verify this, we compared the results of stylization on 3D models obtained with and without CLIP supervision in the first sketch-to-3D model process. Fig. \ref{fig:ablation2} illustrates the difference in texturing accuracy between the two cases. 
\begin{figure*}[!h]
	\centering
	\includegraphics[width=1\textwidth]{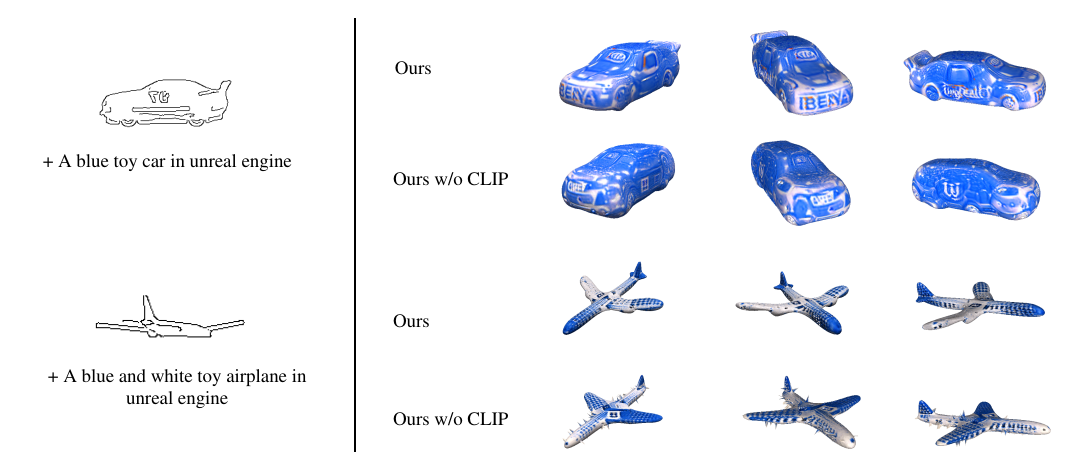}
	\caption{The visualization of Ablation Study. By removing CLIP supervision to the first sketch-to-model process, the later stylization will also be affected, as evident in the inaccurate texture of wheel position in the car and the unwanted spike structure in the airplane.}
	\label{fig:ablation2}
\end{figure*}The models obtained in the sketch-to-model process without CLIP supervision exhibit less accurate and less realistic features, confirming the effectiveness of CLIP supervision in generating more realistic features. The stylization results for models w/o CLIP supervision at sketch-to-model process are unsatisfactory.
The aligned features enabled by CLIP supervision can improve the quality of the CLIP-enabled stylization post-processing.

\subsection{Limitations and Future Directions}
{\color{black}
Despite generating satisfactory results, our method has certain limitations. Firstly, since our method predicts the offset from a template mesh, it may encounter difficulties when dealing with shapes that have a non-zero genus. Secondly, as humans possess diverse drawing capabilities, our method's deterministic regression approach may reflect the drawing skill (or lack thereof) in the final result. If the input sketch is too abstract or ambiguous, the method may fail to produce high-quality results. In such cases, a conditioned generation framework might be more effective \cite{chen2024rapid}, but further research is needed to explore this possibility.

Fig. \ref{fig:fig_failure} illustrates some instances where our method fails. It is evident that the network is misled by asymmetric inputs (shown in the first row) and abstract drawings (shown in the second row), resulting in inaccurate outputs.
}

\begin{figure}[ht]
\centering
\includegraphics[width=0.75\textwidth]{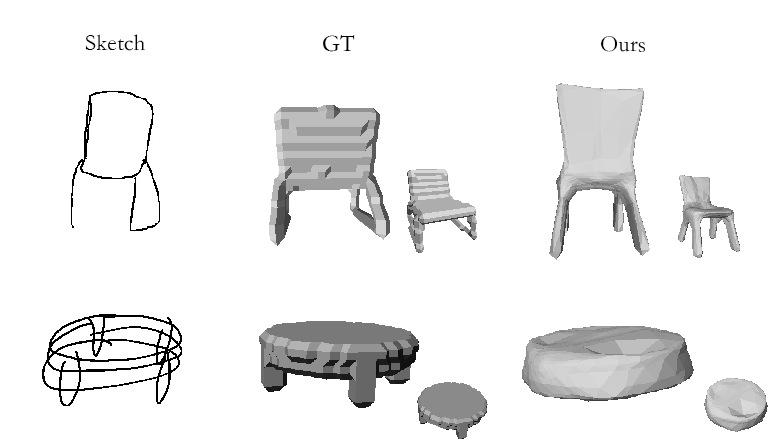}
\caption{
{\color{black}
Some Failure Cases of Our Method on Real Hand-Drawn Sketches}}
\label{fig:fig_failure}
\end{figure}

\section{Conclusion}
\label{sec5}
In this study, we introduced a novel 3D modeling approach \textbf{Magic3D-Sketch}, a framework for fast and intuitive 3D model creation. On the contrary to skill-demanding CAD based 3D modeling, our approach leverages the power of intuitive sketching and text prompts, and propose to use language-image pretraining for extra supervision. We introduce a neural network that incorporate CLIP model to generate high-fidelity 3D models with fine details. We also propose an approach to add color to the sketch-derived 3D models by leveraging the features obtained from the CLIP model, which is, to the best of our knowledge, the first in this field. Our experiments showed that our method outperformed existing state-of-the-art methods in terms of both quantitative metrics and user ratings. We believe that our work opens up new possibilities for fast and intuitive 3D modeling and has significant potential for applications in various fields, such as architecture, game development, and art design.



\section*{Declaration of competing interest}
Tianrun Chen and Chaotao Ding works for KOKONI 3D, a 3D printing company in China. Global patents of this research work has been filed by KOKONI3D, Moxin (Huzhou) Technology Co., LTD.

\section*{Acknowledgments}
This paper is supported by the Public Welfare Research Program of Huzhou Science and Technology Bureau (2022GZ01) and Moxin (Huzhou) Technology Co., LTD.

\bibliographystyle{model1-num-names}
\bibliography{sample-base}

\begin{thebibliography}{59}
\expandafter\ifx\csname natexlab\endcsname\relax\def\natexlab#1{#1}\fi
\providecommand{\url}[1]{\texttt{#1}}
\providecommand{\href}[2]{#2}
\providecommand{\path}[1]{#1}
\providecommand{\DOIprefix}{doi:}
\providecommand{\ArXivprefix}{arXiv:}
\providecommand{\URLprefix}{URL: }
\providecommand{\Pubmedprefix}{pmid:}
\providecommand{\doi}[1]{\href{http://dx.doi.org/#1}{\path{#1}}}
\providecommand{\Pubmed}[1]{\href{pmid:#1}{\path{#1}}}
\providecommand{\bibinfo}[2]{#2}
\ifx\xfnm\relax \def\xfnm[#1]{\unskip,\space#1}\fi
\bibitem[{Wang et~al.(2020)Wang, Lyu, Li, and Zhang}]{wang2020vr}
\bibinfo{author}{M.~Wang}, \bibinfo{author}{X.-Q. Lyu}, \bibinfo{author}{Y.-J. Li}, \bibinfo{author}{F.-L. Zhang},
\newblock \bibinfo{title}{Vr content creation and exploration with deep learning: A survey},
\newblock \bibinfo{journal}{Computational Visual Media} \bibinfo{volume}{6} (\bibinfo{year}{2020}) \bibinfo{pages}{3--28}.
\bibitem[{Cohen et~al.(1999)Cohen, Markosian, Zeleznik, Hughes, and Barzel}]{cohen1999interface}
\bibinfo{author}{J.~M. Cohen}, \bibinfo{author}{L.~Markosian}, \bibinfo{author}{R.~C. Zeleznik}, \bibinfo{author}{J.~F. Hughes}, \bibinfo{author}{R.~Barzel},
\newblock \bibinfo{title}{An interface for sketching 3d curves},
\newblock in: \bibinfo{booktitle}{Proceedings of the 1999 symposium on Interactive 3D graphics}, \bibinfo{year}{1999}, pp. \bibinfo{pages}{17--21}.
\bibitem[{Ramesh et~al.(2021)Ramesh, Pavlov, Goh, Gray, Voss, Radford, Chen, and Sutskever}]{ramesh2021zero}
\bibinfo{author}{A.~Ramesh}, \bibinfo{author}{M.~Pavlov}, \bibinfo{author}{G.~Goh}, \bibinfo{author}{S.~Gray}, \bibinfo{author}{C.~Voss}, \bibinfo{author}{A.~Radford}, \bibinfo{author}{M.~Chen}, \bibinfo{author}{I.~Sutskever},
\newblock \bibinfo{title}{Zero-shot text-to-image generation},
\newblock in: \bibinfo{booktitle}{International Conference on Machine Learning}, \bibinfo{organization}{PMLR}, \bibinfo{year}{2021}, pp. \bibinfo{pages}{8821--8831}.
\bibitem[{Lin et~al.(2022)Lin, Gao, Tang, Takikawa, Zeng, Huang, Kreis, Fidler, Liu, and Lin}]{lin2022magic3d}
\bibinfo{author}{C.-H. Lin}, \bibinfo{author}{J.~Gao}, \bibinfo{author}{L.~Tang}, \bibinfo{author}{T.~Takikawa}, \bibinfo{author}{X.~Zeng}, \bibinfo{author}{X.~Huang}, \bibinfo{author}{K.~Kreis}, \bibinfo{author}{S.~Fidler}, \bibinfo{author}{M.-Y. Liu}, \bibinfo{author}{T.-Y. Lin},
\newblock \bibinfo{title}{Magic3d: High-resolution text-to-3d content creation},
\newblock \bibinfo{journal}{arXiv preprint arXiv:2211.10440}  (\bibinfo{year}{2022}).
\bibitem[{Deng et~al.(2020)Deng, Huang, and Yang}]{deng2020interactive}
\bibinfo{author}{C.~Deng}, \bibinfo{author}{J.~Huang}, \bibinfo{author}{Y.-L. Yang},
\newblock \bibinfo{title}{Interactive modeling of lofted shapes from a single image},
\newblock \bibinfo{journal}{Computational Visual Media} \bibinfo{volume}{6} (\bibinfo{year}{2020}) \bibinfo{pages}{279--289}.
\bibitem[{Zhang et~al.(2021)Zhang, Guo, and Gu}]{zhang2021sketch2model}
\bibinfo{author}{S.-H. Zhang}, \bibinfo{author}{Y.-C. Guo}, \bibinfo{author}{Q.-W. Gu},
\newblock \bibinfo{title}{Sketch2model: View-aware 3d modeling from single free-hand sketches},
\newblock in: \bibinfo{booktitle}{Proceedings of the IEEE/CVF Conference on Computer Vision and Pattern Recognition (CVPR)}, \bibinfo{year}{2021}, pp. \bibinfo{pages}{6012--6021}.
\bibitem[{Guillard et~al.(2021)Guillard, Remelli, Yvernay, and Fua}]{guillard2021sketch2mesh}
\bibinfo{author}{B.~Guillard}, \bibinfo{author}{E.~Remelli}, \bibinfo{author}{P.~Yvernay}, \bibinfo{author}{P.~Fua},
\newblock \bibinfo{title}{Sketch2mesh: Reconstructing and editing 3d shapes from sketches},
\newblock in: \bibinfo{booktitle}{Proceedings of the IEEE/CVF International Conference on Computer Vision (ICCV)}, \bibinfo{year}{2021}, pp. \bibinfo{pages}{13023--13032}.
\bibitem[{Brown et~al.(2020)Brown, Mann, Ryder, Subbiah, Kaplan, Dhariwal, Neelakantan, Shyam, Sastry, Askell et~al.}]{brown2020gpt}
\bibinfo{author}{T.~Brown}, \bibinfo{author}{B.~Mann}, \bibinfo{author}{N.~Ryder}, \bibinfo{author}{M.~Subbiah}, \bibinfo{author}{J.~D. Kaplan}, \bibinfo{author}{P.~Dhariwal}, \bibinfo{author}{A.~Neelakantan}, \bibinfo{author}{P.~Shyam}, \bibinfo{author}{G.~Sastry}, \bibinfo{author}{A.~Askell}, et~al.,
\newblock \bibinfo{title}{Language models are few-shot learners},
\newblock \bibinfo{journal}{Advances in neural information processing systems} \bibinfo{volume}{33} (\bibinfo{year}{2020}) \bibinfo{pages}{1877--1901}.
\bibitem[{Betker et~al.(2023)Betker, Goh, Jing, Brooks, Wang, Li, Ouyang, Zhuang, Lee, Guo et~al.}]{betker2023dalle}
\bibinfo{author}{J.~Betker}, \bibinfo{author}{G.~Goh}, \bibinfo{author}{L.~Jing}, \bibinfo{author}{T.~Brooks}, \bibinfo{author}{J.~Wang}, \bibinfo{author}{L.~Li}, \bibinfo{author}{L.~Ouyang}, \bibinfo{author}{J.~Zhuang}, \bibinfo{author}{J.~Lee}, \bibinfo{author}{Y.~Guo}, et~al.,
\newblock \bibinfo{title}{Improving image generation with better captions},
\newblock \bibinfo{journal}{Computer Science. https://cdn. openai. com/papers/dall-e-3. pdf} \bibinfo{volume}{2} (\bibinfo{year}{2023}) \bibinfo{pages}{3}.
\bibitem[{Kirillov et~al.(2023)Kirillov, Mintun, Ravi, Mao, Rolland, Gustafson, Xiao, Whitehead, Berg, Lo et~al.}]{kirillov2023segment}
\bibinfo{author}{A.~Kirillov}, \bibinfo{author}{E.~Mintun}, \bibinfo{author}{N.~Ravi}, \bibinfo{author}{H.~Mao}, \bibinfo{author}{C.~Rolland}, \bibinfo{author}{L.~Gustafson}, \bibinfo{author}{T.~Xiao}, \bibinfo{author}{S.~Whitehead}, \bibinfo{author}{A.~C. Berg}, \bibinfo{author}{W.-Y. Lo}, et~al.,
\newblock \bibinfo{title}{Segment anything},
\newblock \bibinfo{journal}{arXiv preprint arXiv:2304.02643}  (\bibinfo{year}{2023}).
\bibitem[{Bommasani et~al.(2021)Bommasani, Hudson, Adeli, Altman, Arora, von Arx, Bernstein, Bohg, Bosselut, Brunskill et~al.}]{bommasani2021opportunities}
\bibinfo{author}{R.~Bommasani}, \bibinfo{author}{D.~A. Hudson}, \bibinfo{author}{E.~Adeli}, \bibinfo{author}{R.~Altman}, \bibinfo{author}{S.~Arora}, \bibinfo{author}{S.~von Arx}, \bibinfo{author}{M.~S. Bernstein}, \bibinfo{author}{J.~Bohg}, \bibinfo{author}{A.~Bosselut}, \bibinfo{author}{E.~Brunskill}, et~al.,
\newblock \bibinfo{title}{On the opportunities and risks of foundation models},
\newblock \bibinfo{journal}{arXiv preprint arXiv:2108.07258}  (\bibinfo{year}{2021}).
\bibitem[{Radford et~al.(2021)Radford, Kim, Hallacy, Ramesh, Goh, Agarwal, Sastry, Askell, Mishkin, Clark et~al.}]{radford2021learning}
\bibinfo{author}{A.~Radford}, \bibinfo{author}{J.~W. Kim}, \bibinfo{author}{C.~Hallacy}, \bibinfo{author}{A.~Ramesh}, \bibinfo{author}{G.~Goh}, \bibinfo{author}{S.~Agarwal}, \bibinfo{author}{G.~Sastry}, \bibinfo{author}{A.~Askell}, \bibinfo{author}{P.~Mishkin}, \bibinfo{author}{J.~Clark}, et~al.,
\newblock \bibinfo{title}{Learning transferable visual models from natural language supervision},
\newblock in: \bibinfo{booktitle}{International conference on machine learning}, \bibinfo{organization}{PMLR}, \bibinfo{year}{2021}, pp. \bibinfo{pages}{8748--8763}.
\bibitem[{Bonnici et~al.(2019)Bonnici, Akman, Calleja, Camilleri, Fehling, Ferreira, Hermuth, Israel, Landwehr, Liu et~al.}]{bonnici2019sketch}
\bibinfo{author}{A.~Bonnici}, \bibinfo{author}{A.~Akman}, \bibinfo{author}{G.~Calleja}, \bibinfo{author}{K.~P. Camilleri}, \bibinfo{author}{P.~Fehling}, \bibinfo{author}{A.~Ferreira}, \bibinfo{author}{F.~Hermuth}, \bibinfo{author}{J.~H. Israel}, \bibinfo{author}{T.~Landwehr}, \bibinfo{author}{J.~Liu}, et~al.,
\newblock \bibinfo{title}{Sketch-based interaction and modeling: where do we stand?},
\newblock \bibinfo{journal}{AI EDAM} \bibinfo{volume}{33} (\bibinfo{year}{2019}) \bibinfo{pages}{370--388}.
\bibitem[{Olsen et~al.(2009)Olsen, Samavati, Sousa, and Jorge}]{olsen2009sketch}
\bibinfo{author}{L.~Olsen}, \bibinfo{author}{F.~F. Samavati}, \bibinfo{author}{M.~C. Sousa}, \bibinfo{author}{J.~A. Jorge},
\newblock \bibinfo{title}{Sketch-based modeling: A survey},
\newblock \bibinfo{journal}{Computers \& Graphics} \bibinfo{volume}{33} (\bibinfo{year}{2009}) \bibinfo{pages}{85--103}.
\bibitem[{Li et~al.(2020)Li, Pan, Bousseau, and Mitra}]{li2020sketch2cad}
\bibinfo{author}{C.~Li}, \bibinfo{author}{H.~Pan}, \bibinfo{author}{A.~Bousseau}, \bibinfo{author}{N.~J. Mitra},
\newblock \bibinfo{title}{Sketch2cad: Sequential cad modeling by sketching in context},
\newblock \bibinfo{journal}{ACM Transactions on Graphics (TOG)} \bibinfo{volume}{39} (\bibinfo{year}{2020}) \bibinfo{pages}{1--14}.
\bibitem[{Shtof et~al.(2013)Shtof, Agathos, Gingold, Shamir, and Cohen-Or}]{shtof2013geosemantic}
\bibinfo{author}{A.~Shtof}, \bibinfo{author}{A.~Agathos}, \bibinfo{author}{Y.~Gingold}, \bibinfo{author}{A.~Shamir}, \bibinfo{author}{D.~Cohen-Or},
\newblock \bibinfo{title}{Geosemantic snapping for sketch-based modeling},
\newblock in: \bibinfo{booktitle}{Computer graphics forum}, volume~\bibinfo{volume}{32}, \bibinfo{organization}{Wiley Online Library}, \bibinfo{year}{2013}, pp. \bibinfo{pages}{245--253}.
\bibitem[{Chen et~al.(2003)Chen, Tian, Shen, and Ouhyoung}]{chen2003visual}
\bibinfo{author}{D.-Y. Chen}, \bibinfo{author}{X.-P. Tian}, \bibinfo{author}{Y.-T. Shen}, \bibinfo{author}{M.~Ouhyoung},
\newblock \bibinfo{title}{On visual similarity based 3d model retrieval},
\newblock in: \bibinfo{booktitle}{Computer graphics forum}, volume~\bibinfo{volume}{22}, \bibinfo{organization}{Wiley Online Library}, \bibinfo{year}{2003}, pp. \bibinfo{pages}{223--232}.
\bibitem[{Wang et~al.(2015)Wang, Kang, and Li}]{wang2015sketch}
\bibinfo{author}{F.~Wang}, \bibinfo{author}{L.~Kang}, \bibinfo{author}{Y.~Li},
\newblock \bibinfo{title}{Sketch-based 3d shape retrieval using convolutional neural networks},
\newblock in: \bibinfo{booktitle}{Proceedings of the IEEE conference on computer vision and pattern recognition}, \bibinfo{year}{2015}, pp. \bibinfo{pages}{1875--1883}.
\bibitem[{Sangkloy et~al.(2016)Sangkloy, Burnell, Ham, and Hays}]{sangkloy2016sketchy}
\bibinfo{author}{P.~Sangkloy}, \bibinfo{author}{N.~Burnell}, \bibinfo{author}{C.~Ham}, \bibinfo{author}{J.~Hays},
\newblock \bibinfo{title}{The sketchy database: learning to retrieve badly drawn bunnies},
\newblock \bibinfo{journal}{ACM Transactions on Graphics (TOG)} \bibinfo{volume}{35} (\bibinfo{year}{2016}) \bibinfo{pages}{1--12}.
\bibitem[{Xu et~al.(2020)Xu, Hu, Wattanachote, Zeng, and Gong}]{8960471}
\bibinfo{author}{Y.~Xu}, \bibinfo{author}{J.~Hu}, \bibinfo{author}{K.~Wattanachote}, \bibinfo{author}{K.~Zeng}, \bibinfo{author}{Y.~Gong},
\newblock \bibinfo{title}{Sketch-based shape retrieval via best view selection and a cross-domain similarity measure},
\newblock \bibinfo{journal}{IEEE Transactions on Multimedia} \bibinfo{volume}{22} (\bibinfo{year}{2020}) \bibinfo{pages}{2950--2962}.
\bibitem[{Yuan et~al.(2023)Yuan, Wen, Liu, and Fang}]{10155453}
\bibinfo{author}{S.~Yuan}, \bibinfo{author}{C.~Wen}, \bibinfo{author}{Y.-S. Liu}, \bibinfo{author}{Y.~Fang},
\newblock \bibinfo{title}{Retrieval-specific view learning for sketch-to-shape retrieval},
\newblock \bibinfo{journal}{IEEE Transactions on Multimedia}  (\bibinfo{year}{2023}) \bibinfo{pages}{1--12}.
\bibitem[{Nie et~al.(2020)Nie, Ren, Liu, Mao, and Nie}]{nie2020m}
\bibinfo{author}{W.-Z. Nie}, \bibinfo{author}{M.-J. Ren}, \bibinfo{author}{A.-A. Liu}, \bibinfo{author}{Z.~Mao}, \bibinfo{author}{J.~Nie},
\newblock \bibinfo{title}{M-gcn: Multi-branch graph convolution network for 2d image-based on 3d model retrieval},
\newblock \bibinfo{journal}{IEEE Transactions on Multimedia} \bibinfo{volume}{23} (\bibinfo{year}{2020}) \bibinfo{pages}{1962--1976}.
\bibitem[{Chen et~al.(2023{\natexlab{a}})Chen, Fu, Zang, Zhu, Zhang, Mao, and Sun}]{chen2023deep3dsketch+}
\bibinfo{author}{T.~Chen}, \bibinfo{author}{C.~Fu}, \bibinfo{author}{Y.~Zang}, \bibinfo{author}{L.~Zhu}, \bibinfo{author}{J.~Zhang}, \bibinfo{author}{P.~Mao}, \bibinfo{author}{L.~Sun},
\newblock \bibinfo{title}{Deep3dsketch+: Rapid 3d modeling from single free-hand sketches},
\newblock in: \bibinfo{booktitle}{MultiMedia Modeling: 29th International Conference, MMM 2023, Bergen, Norway, January 9--12, 2023, Proceedings, Part II}, \bibinfo{organization}{Springer}, \bibinfo{year}{2023}{\natexlab{a}}, pp. \bibinfo{pages}{16--28}.
\bibitem[{Chen et~al.(2023{\natexlab{b}})Chen, Ding, Zhu, Zang, Liao, Li, and Sun}]{chen2023reality3dsketch}
\bibinfo{author}{T.~Chen}, \bibinfo{author}{C.~Ding}, \bibinfo{author}{L.~Zhu}, \bibinfo{author}{Y.~Zang}, \bibinfo{author}{Y.~Liao}, \bibinfo{author}{Z.~Li}, \bibinfo{author}{L.~Sun},
\newblock \bibinfo{title}{Reality3dsketch: rapid 3d modeling of objects from single freehand sketches},
\newblock \bibinfo{journal}{IEEE Transactions on Multimedia}  (\bibinfo{year}{2023}{\natexlab{b}}).
\bibitem[{Zang et~al.(2023)Zang, Fu, Chen, Hu, Liu, and Hu}]{zang2023deep3dsketch+}
\bibinfo{author}{Y.~Zang}, \bibinfo{author}{C.~Fu}, \bibinfo{author}{T.~Chen}, \bibinfo{author}{Y.~Hu}, \bibinfo{author}{Q.~Liu}, \bibinfo{author}{W.~Hu},
\newblock \bibinfo{title}{Deep3dsketch+: obtaining customized 3d model by single free-hand sketch through deep learning},
\newblock \bibinfo{journal}{arXiv preprint arXiv:2310.18609}  (\bibinfo{year}{2023}).
\bibitem[{Chang et~al.(2015)Chang, Funkhouser, Guibas, Hanrahan, Huang, Li, Savarese, Savva, Song, Su et~al.}]{chang2015shapenet}
\bibinfo{author}{A.~X. Chang}, \bibinfo{author}{T.~Funkhouser}, \bibinfo{author}{L.~Guibas}, \bibinfo{author}{P.~Hanrahan}, \bibinfo{author}{Q.~Huang}, \bibinfo{author}{Z.~Li}, \bibinfo{author}{S.~Savarese}, \bibinfo{author}{M.~Savva}, \bibinfo{author}{S.~Song}, \bibinfo{author}{H.~Su}, et~al.,
\newblock \bibinfo{title}{Shapenet: An information-rich 3d model repository},
\newblock \bibinfo{journal}{arXiv preprint arXiv:1512.03012}  (\bibinfo{year}{2015}).
\bibitem[{Chen and Zhang(2019)}]{chen2019learning}
\bibinfo{author}{Z.~Chen}, \bibinfo{author}{H.~Zhang},
\newblock \bibinfo{title}{Learning implicit fields for generative shape modeling},
\newblock in: \bibinfo{booktitle}{Proceedings of the IEEE/CVF Conference on Computer Vision and Pattern Recognition (CVPR)}, \bibinfo{year}{2019}, pp. \bibinfo{pages}{5939--5948}.
\bibitem[{Park et~al.(2019)Park, Florence, Straub, Newcombe, and Lovegrove}]{park2019deepsdf}
\bibinfo{author}{J.~J. Park}, \bibinfo{author}{P.~Florence}, \bibinfo{author}{J.~Straub}, \bibinfo{author}{R.~Newcombe}, \bibinfo{author}{S.~Lovegrove},
\newblock \bibinfo{title}{Deepsdf: Learning continuous signed distance functions for shape representation},
\newblock in: \bibinfo{booktitle}{Proceedings of the IEEE/CVF conference on computer vision and pattern recognition (CVPR)}, \bibinfo{year}{2019}, pp. \bibinfo{pages}{165--174}.
\bibitem[{Liu et~al.(2019{\natexlab{a}})Liu, Li, Chen, and Li}]{liu2019soft}
\bibinfo{author}{S.~Liu}, \bibinfo{author}{T.~Li}, \bibinfo{author}{W.~Chen}, \bibinfo{author}{H.~Li},
\newblock \bibinfo{title}{Soft rasterizer: A differentiable renderer for image-based 3d reasoning},
\newblock in: \bibinfo{booktitle}{Proceedings of the IEEE/CVF International Conference on Computer Vision (ICCV)}, \bibinfo{year}{2019}{\natexlab{a}}, pp. \bibinfo{pages}{7708--7717}.
\bibitem[{Liu et~al.(2019{\natexlab{b}})Liu, Saito, Chen, and Li}]{liu2019learning}
\bibinfo{author}{S.~Liu}, \bibinfo{author}{S.~Saito}, \bibinfo{author}{W.~Chen}, \bibinfo{author}{H.~Li},
\newblock \bibinfo{title}{Learning to infer implicit surfaces without 3d supervision},
\newblock \bibinfo{journal}{Advances in Neural Information Processing Systems (NeurIPS)} \bibinfo{volume}{32} (\bibinfo{year}{2019}{\natexlab{b}}).
\bibitem[{Kato et~al.(2018)Kato, Ushiku, and Harada}]{kato2018neural}
\bibinfo{author}{H.~Kato}, \bibinfo{author}{Y.~Ushiku}, \bibinfo{author}{T.~Harada},
\newblock \bibinfo{title}{Neural 3d mesh renderer},
\newblock in: \bibinfo{booktitle}{Proceedings of the IEEE conference on computer vision and pattern recognition}, \bibinfo{year}{2018}, pp. \bibinfo{pages}{3907--3916}.
\bibitem[{Gao et~al.(2022)Gao, Kong, Wang, Li, and Yin}]{9817616}
\bibinfo{author}{J.~Gao}, \bibinfo{author}{D.~Kong}, \bibinfo{author}{S.~Wang}, \bibinfo{author}{J.~Li}, \bibinfo{author}{B.~Yin},
\newblock \bibinfo{title}{Dasi: Learning domain adaptive shape impression for 3d object reconstruction},
\newblock \bibinfo{journal}{IEEE Transactions on Multimedia}  (\bibinfo{year}{2022}) \bibinfo{pages}{1--15}.
\bibitem[{Lin et~al.(2020)Lin, Wang, and Lucey}]{lin2020sdf}
\bibinfo{author}{C.-H. Lin}, \bibinfo{author}{C.~Wang}, \bibinfo{author}{S.~Lucey},
\newblock \bibinfo{title}{Sdf-srn: Learning signed distance 3d object reconstruction from static images},
\newblock \bibinfo{journal}{Advances in Neural Information Processing Systems (NeurIPS)} \bibinfo{volume}{33} (\bibinfo{year}{2020}) \bibinfo{pages}{11453--11464}.
\bibitem[{Yu et~al.(2021)Yu, Ye, Tancik, and Kanazawa}]{yu2021pixelnerf}
\bibinfo{author}{A.~Yu}, \bibinfo{author}{V.~Ye}, \bibinfo{author}{M.~Tancik}, \bibinfo{author}{A.~Kanazawa},
\newblock \bibinfo{title}{pixelnerf: Neural radiance fields from one or few images},
\newblock in: \bibinfo{booktitle}{Proceedings of the IEEE/CVF Conference on Computer Vision and Pattern Recognition (CVPR)}, \bibinfo{year}{2021}, pp. \bibinfo{pages}{4578--4587}.
\bibitem[{Jain et~al.(2022)Jain, Mildenhall, Barron, Abbeel, and Poole}]{jain2022zero}
\bibinfo{author}{A.~Jain}, \bibinfo{author}{B.~Mildenhall}, \bibinfo{author}{J.~T. Barron}, \bibinfo{author}{P.~Abbeel}, \bibinfo{author}{B.~Poole},
\newblock \bibinfo{title}{Zero-shot text-guided object generation with dream fields},
\newblock in: \bibinfo{booktitle}{Proceedings of the IEEE/CVF Conference on Computer Vision and Pattern Recognition}, \bibinfo{year}{2022}, pp. \bibinfo{pages}{867--876}.
\bibitem[{Mohammad~Khalid et~al.(2022)Mohammad~Khalid, Xie, Belilovsky, and Popa}]{mohammad2022clip}
\bibinfo{author}{N.~Mohammad~Khalid}, \bibinfo{author}{T.~Xie}, \bibinfo{author}{E.~Belilovsky}, \bibinfo{author}{T.~Popa},
\newblock \bibinfo{title}{Clip-mesh: Generating textured meshes from text using pretrained image-text models},
\newblock in: \bibinfo{booktitle}{SIGGRAPH Asia 2022 conference papers}, \bibinfo{year}{2022}, pp. \bibinfo{pages}{1--8}.
\bibitem[{Xu et~al.(2023)Xu, Wang, Cheng, Cao, Shan, Qie, and Gao}]{xu2023dream3d}
\bibinfo{author}{J.~Xu}, \bibinfo{author}{X.~Wang}, \bibinfo{author}{W.~Cheng}, \bibinfo{author}{Y.-P. Cao}, \bibinfo{author}{Y.~Shan}, \bibinfo{author}{X.~Qie}, \bibinfo{author}{S.~Gao},
\newblock \bibinfo{title}{Dream3d: Zero-shot text-to-3d synthesis using 3d shape prior and text-to-image diffusion models},
\newblock in: \bibinfo{booktitle}{Proceedings of the IEEE/CVF Conference on Computer Vision and Pattern Recognition}, \bibinfo{year}{2023}, pp. \bibinfo{pages}{20908--20918}.
\bibitem[{Jetchev(2021)}]{jetchev2021clipmatrix}
\bibinfo{author}{N.~Jetchev},
\newblock \bibinfo{title}{Clipmatrix: Text-controlled creation of 3d textured meshes},
\newblock \bibinfo{journal}{arXiv preprint arXiv:2109.12922}  (\bibinfo{year}{2021}).
\bibitem[{Canfes et~al.(2023)Canfes, Atasoy, Dirik, and Yanardag}]{canfes2023text}
\bibinfo{author}{Z.~Canfes}, \bibinfo{author}{M.~F. Atasoy}, \bibinfo{author}{A.~Dirik}, \bibinfo{author}{P.~Yanardag},
\newblock \bibinfo{title}{Text and image guided 3d avatar generation and manipulation},
\newblock in: \bibinfo{booktitle}{Proceedings of the IEEE/CVF Winter Conference on Applications of Computer Vision}, \bibinfo{year}{2023}, pp. \bibinfo{pages}{4421--4431}.
\bibitem[{Lee and Chang(2022)}]{lee2022understanding}
\bibinfo{author}{H.-H. Lee}, \bibinfo{author}{A.~X. Chang},
\newblock \bibinfo{title}{Understanding pure clip guidance for voxel grid nerf models},
\newblock \bibinfo{journal}{arXiv preprint arXiv:2209.15172}  (\bibinfo{year}{2022}).
\bibitem[{Hong et~al.(2022)Hong, Zhang, Pan, Cai, Yang, and Liu}]{hong2022avatarclip}
\bibinfo{author}{F.~Hong}, \bibinfo{author}{M.~Zhang}, \bibinfo{author}{L.~Pan}, \bibinfo{author}{Z.~Cai}, \bibinfo{author}{L.~Yang}, \bibinfo{author}{Z.~Liu},
\newblock \bibinfo{title}{Avatarclip: Zero-shot text-driven generation and animation of 3d avatars},
\newblock \bibinfo{journal}{arXiv preprint arXiv:2205.08535}  (\bibinfo{year}{2022}).
\bibitem[{Sanghi et~al.(2022)Sanghi, Chu, Lambourne, Wang, Cheng, Fumero, and Malekshan}]{sanghi2022clip}
\bibinfo{author}{A.~Sanghi}, \bibinfo{author}{H.~Chu}, \bibinfo{author}{J.~G. Lambourne}, \bibinfo{author}{Y.~Wang}, \bibinfo{author}{C.-Y. Cheng}, \bibinfo{author}{M.~Fumero}, \bibinfo{author}{K.~R. Malekshan},
\newblock \bibinfo{title}{Clip-forge: Towards zero-shot text-to-shape generation},
\newblock in: \bibinfo{booktitle}{Proceedings of the IEEE/CVF Conference on Computer Vision and Pattern Recognition}, \bibinfo{year}{2022}, pp. \bibinfo{pages}{18603--18613}.
\bibitem[{Liu et~al.(2023)Liu, Dai, Li, Qi, and Fu}]{liu2023iss++}
\bibinfo{author}{Z.~Liu}, \bibinfo{author}{P.~Dai}, \bibinfo{author}{R.~Li}, \bibinfo{author}{X.~Qi}, \bibinfo{author}{C.-W. Fu},
\newblock \bibinfo{title}{Iss++: Image as stepping stone for text-guided 3d shape generation},
\newblock \bibinfo{journal}{arXiv preprint arXiv:2303.15181}  (\bibinfo{year}{2023}).
\bibitem[{Chen et~al.(2023)Chen, Siddiqui, Lee, Tulyakov, and Nie{\ss}ner}]{chen2023text2tex}
\bibinfo{author}{D.~Z. Chen}, \bibinfo{author}{Y.~Siddiqui}, \bibinfo{author}{H.-Y. Lee}, \bibinfo{author}{S.~Tulyakov}, \bibinfo{author}{M.~Nie{\ss}ner},
\newblock \bibinfo{title}{Text2tex: Text-driven texture synthesis via diffusion models},
\newblock \bibinfo{journal}{arXiv preprint arXiv:2303.11396}  (\bibinfo{year}{2023}).
\bibitem[{Michel et~al.(2022)Michel, Bar-On, Liu, Benaim, and Hanocka}]{michel2022text2mesh}
\bibinfo{author}{O.~Michel}, \bibinfo{author}{R.~Bar-On}, \bibinfo{author}{R.~Liu}, \bibinfo{author}{S.~Benaim}, \bibinfo{author}{R.~Hanocka},
\newblock \bibinfo{title}{Text2mesh: Text-driven neural stylization for meshes},
\newblock in: \bibinfo{booktitle}{Proceedings of the IEEE/CVF Conference on Computer Vision and Pattern Recognition}, \bibinfo{year}{2022}, pp. \bibinfo{pages}{13492--13502}.
\bibitem[{Wei et~al.(2023)Wei, Wang, Feng, Lin, and Yap}]{wei2023taps3d}
\bibinfo{author}{J.~Wei}, \bibinfo{author}{H.~Wang}, \bibinfo{author}{J.~Feng}, \bibinfo{author}{G.~Lin}, \bibinfo{author}{K.-H. Yap},
\newblock \bibinfo{title}{Taps3d: Text-guided 3d textured shape generation from pseudo supervision},
\newblock in: \bibinfo{booktitle}{Proceedings of the IEEE/CVF Conference on Computer Vision and Pattern Recognition}, \bibinfo{year}{2023}, pp. \bibinfo{pages}{16805--16815}.
\bibitem[{Xu et~al.(2019)Xu, Wang, Ceylan, Mech, and Neumann}]{xu2019disn}
\bibinfo{author}{Q.~Xu}, \bibinfo{author}{W.~Wang}, \bibinfo{author}{D.~Ceylan}, \bibinfo{author}{R.~Mech}, \bibinfo{author}{U.~Neumann},
\newblock \bibinfo{title}{Disn: Deep implicit surface network for high-quality single-view 3d reconstruction},
\newblock \bibinfo{journal}{Advances in neural information processing systems} \bibinfo{volume}{32} (\bibinfo{year}{2019}).
\bibitem[{Chen et~al.(2024)Chen, Cao, Li, Zang, and Sun}]{chen2024deep3dsketchim}
\bibinfo{author}{T.~Chen}, \bibinfo{author}{R.~Cao}, \bibinfo{author}{Z.~Li}, \bibinfo{author}{Y.~Zang}, \bibinfo{author}{L.~Sun},
\newblock \bibinfo{title}{Deep3dsketch-im: rapid high-fidelity ai 3d model generation by single freehand sketches},
\newblock \bibinfo{journal}{Frontiers of Information Technology \& Electronic Engineering} \bibinfo{volume}{25} (\bibinfo{year}{2024}) \bibinfo{pages}{149--159}.
\bibitem[{Gadelha et~al.(2019)Gadelha, Wang, and Maji}]{gadelha2019shape}
\bibinfo{author}{M.~Gadelha}, \bibinfo{author}{R.~Wang}, \bibinfo{author}{S.~Maji},
\newblock \bibinfo{title}{Shape reconstruction using differentiable projections and deep priors},
\newblock in: \bibinfo{booktitle}{Proceedings of the IEEE/CVF International Conference on Computer Vision (ICCV)}, \bibinfo{year}{2019}, pp. \bibinfo{pages}{22--30}.
\bibitem[{Hu et~al.(2018)Hu, Zhu, Liu, Xie, Tang, Wang, Shen, and Shao}]{hu2018structure}
\bibinfo{author}{X.~Hu}, \bibinfo{author}{F.~Zhu}, \bibinfo{author}{L.~Liu}, \bibinfo{author}{J.~Xie}, \bibinfo{author}{J.~Tang}, \bibinfo{author}{N.~Wang}, \bibinfo{author}{F.~Shen}, \bibinfo{author}{L.~Shao},
\newblock \bibinfo{title}{Structure-aware 3d shape synthesis from single-view images.},
\newblock in: \bibinfo{booktitle}{BMVC}, \bibinfo{year}{2018}, pp. \bibinfo{pages}{230--243}.
\bibitem[{He et~al.(2016)He, Zhang, Ren, and Sun}]{he2016deep}
\bibinfo{author}{K.~He}, \bibinfo{author}{X.~Zhang}, \bibinfo{author}{S.~Ren}, \bibinfo{author}{J.~Sun},
\newblock \bibinfo{title}{Deep residual learning for image recognition},
\newblock in: \bibinfo{booktitle}{Proceedings of the IEEE conference on computer vision and pattern recognition}, \bibinfo{year}{2016}, pp. \bibinfo{pages}{770--778}.
\bibitem[{Kar et~al.(2017)Kar, H{\"a}ne, and Malik}]{kar2017learning}
\bibinfo{author}{A.~Kar}, \bibinfo{author}{C.~H{\"a}ne}, \bibinfo{author}{J.~Malik},
\newblock \bibinfo{title}{Learning a multi-view stereo machine},
\newblock \bibinfo{journal}{Advances in neural information processing systems (NeurIPS)} \bibinfo{volume}{30} (\bibinfo{year}{2017}).
\bibitem[{Cai et~al.(2021)Cai, Wang, Zhu, Cham, Cai, Yuan, Liu, Zheng, Yan, Ding et~al.}]{cai2021unified}
\bibinfo{author}{Y.~Cai}, \bibinfo{author}{Y.~Wang}, \bibinfo{author}{Y.~Zhu}, \bibinfo{author}{T.-J. Cham}, \bibinfo{author}{J.~Cai}, \bibinfo{author}{J.~Yuan}, \bibinfo{author}{J.~Liu}, \bibinfo{author}{C.~Zheng}, \bibinfo{author}{S.~Yan}, \bibinfo{author}{H.~Ding}, et~al.,
\newblock \bibinfo{title}{A unified 3d human motion synthesis model via conditional variational auto-encoder},
\newblock in: \bibinfo{booktitle}{Proceedings of the IEEE/CVF International Conference on Computer Vision (ICCV)}, \bibinfo{year}{2021}, pp. \bibinfo{pages}{11645--11655}.
\bibitem[{Yao et~al.(2022)Yao, Zhong, Yan, Zhai, and Yang}]{yao2022dfa}
\bibinfo{author}{S.~Yao}, \bibinfo{author}{R.~Zhong}, \bibinfo{author}{Y.~Yan}, \bibinfo{author}{G.~Zhai}, \bibinfo{author}{X.~Yang},
\newblock \bibinfo{title}{Dfa-nerf: Personalized talking head generation via disentangled face attributes neural rendering},
\newblock \bibinfo{journal}{arXiv preprint arXiv:2201.00791}  (\bibinfo{year}{2022}).
\bibitem[{Zhang et~al.(2023)Zhang, Peng, Chen, Mou, Lin, Yu, Liao, and Zhou}]{zhang2023painting}
\bibinfo{author}{S.~Zhang}, \bibinfo{author}{S.~Peng}, \bibinfo{author}{T.~Chen}, \bibinfo{author}{L.~Mou}, \bibinfo{author}{H.~Lin}, \bibinfo{author}{K.~Yu}, \bibinfo{author}{Y.~Liao}, \bibinfo{author}{X.~Zhou},
\newblock \bibinfo{title}{Painting 3d nature in 2d: View synthesis of natural scenes from a single semantic mask},
\newblock \bibinfo{journal}{arXiv preprint arXiv:2302.07224}  (\bibinfo{year}{2023}).
\bibitem[{Seufert(2019)}]{seufert2019fundamental}
\bibinfo{author}{M.~Seufert},
\newblock \bibinfo{title}{Fundamental advantages of considering quality of experience distributions over mean opinion scores},
\newblock in: \bibinfo{booktitle}{2019 Eleventh international conference on quality of multimedia experience (QoMEX)}, \bibinfo{organization}{IEEE}, \bibinfo{year}{2019}, pp. \bibinfo{pages}{1--6}.
\bibitem[{Albert and Tullis(2022)}]{albert2022measuring}
\bibinfo{author}{B.~Albert}, \bibinfo{author}{T.~Tullis}, \bibinfo{title}{Measuring the User Experience: Collecting, Analyzing, and Presenting UX Metrics}, \bibinfo{publisher}{Morgan Kaufmann}, \bibinfo{year}{2022}.
\bibitem[{Oh et~al.(2018)Oh, Song, Choi, Kim, Lee, and Suh}]{oh2018lead}
\bibinfo{author}{C.~Oh}, \bibinfo{author}{J.~Song}, \bibinfo{author}{J.~Choi}, \bibinfo{author}{S.~Kim}, \bibinfo{author}{S.~Lee}, \bibinfo{author}{B.~Suh},
\newblock \bibinfo{title}{I lead, you help but only with enough details: Understanding user experience of co-creation with artificial intelligence},
\newblock in: \bibinfo{booktitle}{Proceedings of the 2018 CHI Conference on Human Factors in Computing Systems}, \bibinfo{year}{2018}, pp. \bibinfo{pages}{1--13}.
\bibitem[{Chen et~al.(2024)Chen, Ding, Zhang, Yu, Zang, Li, Peng, and Sun}]{chen2024rapid}
\bibinfo{author}{T.~Chen}, \bibinfo{author}{C.~Ding}, \bibinfo{author}{S.~Zhang}, \bibinfo{author}{C.~Yu}, \bibinfo{author}{Y.~Zang}, \bibinfo{author}{Z.~Li}, \bibinfo{author}{S.~Peng}, \bibinfo{author}{L.~Sun},
\newblock \bibinfo{title}{Rapid 3d model generation with intuitive 3d input},
\newblock in: \bibinfo{booktitle}{Proceedings of the IEEE/CVF Conference on Computer Vision and Pattern Recognition}, \bibinfo{year}{2024}, pp. \bibinfo{pages}{12554--12564}.

\end{thebibliography}

\end{document}